\documentclass{article} 

\usepackage{iclr2027_conference,times}

\iclrfinalcopy

\usepackage[utf8]{inputenc} 
\usepackage[T1]{fontenc}    
\usepackage{hyperref}       
\hypersetup{pdftitle={Beyond the Readout: A Latent Numerical Frame Shared Across Modality and Operation},pdfauthor={Oriol Juan, Alex Ferrando de las Morenas, Pau Rodriguez, Xavier Suau}}
\usepackage{url}            
\usepackage{booktabs}       
\usepackage{amsfonts}       
\usepackage{amssymb}        
\usepackage{nicefrac}       
\usepackage{microtype}      
\usepackage{xcolor}         
\usepackage{amsmath}
\usepackage{mathtools}      
\usepackage{soul}           
\usepackage{longtable}      
\usepackage{arydshln}       
\usepackage{graphicx}
\usepackage{subcaption}
\usepackage{multirow}
\usepackage{xspace}
\usepackage{wrapfig}
\usepackage{cleveref}
\usepackage{algorithm}
\usepackage{algpseudocode}
\usepackage{float}
\usepackage{enumitem}

\usepackage{tikz}
\usetikzlibrary{arrows.meta,positioning,shapes.geometric,fit,backgrounds,calc}

\usepackage{placeins}

\crefname{appendix}{Appendix}{Appendices}
\Crefname{appendix}{Appendix}{Appendices}
\crefname{section}{Section}{Sections}
\Crefname{section}{Section}{Sections}

\usepackage{amsmath,amsfonts,bm}

\def\eqref#1{equation~\ref{#1}}

\def\1{\bm{1}}

\def\vh{{\bm{h}}}

\def\vo{{\bm{o}}}

\def\vq{{\bm{q}}}
\def\vr{{\bm{r}}}

\def\vu{{\bm{u}}}
\def\vv{{\bm{v}}}
\def\vw{{\bm{w}}}

\def\vz{{\bm{z}}}

\def\mA{{\bm{A}}}

\def\mD{{\bm{D}}}

\def\mI{{\bm{I}}}

\def\mL{{\bm{L}}}

\def\mQ{{\bm{Q}}}
\def\mR{{\bm{R}}}

\def\mU{{\bm{U}}}
\def\mV{{\bm{V}}}
\def\mW{{\bm{W}}}

\def\mSigma{{\bm{\Sigma}}}

\DeclareMathAlphabet{\mathsfit}{\encodingdefault}{\sfdefault}{m}{sl}
\SetMathAlphabet{\mathsfit}{bold}{\encodingdefault}{\sfdefault}{bx}{n}

\newcommand{\R}{\mathbb{R}}

\DeclareMathOperator*{\argmin}{arg\,min}

\newcommand{\eg}{\emph{e.g.,}\xspace}

\newcommand{\gemma}{\mbox{Gemma4-12B-it}\xspace}
\newcommand{\llama}{\mbox{Llama3.1-8B}\xspace}
\newcommand{\ministral}{\mbox{Ministral-3-14B}\xspace}
\newcommand{\das}{DAS\xspace}
\newcommand{\iia}{\ensuremath{\mathrm{IIA}}}
\newcommand{\iiaar}{\ensuremath{\mathrm{IIA}_{\mathrm{AR}}}}

\newcommand{\tsc}[2]{\ensuremath{\tau_{#1 \rightarrow #2}}}
\newcommand{\Rrand}{\ensuremath{\mR_{\mathrm{rand}}}}

\DeclareRobustCommand{\smalloplus}{%
  \mathbin{\vcenter{\hbox{%
    \begin{tikzpicture}[baseline]
      \draw (0,0) circle (0.5ex);
      \draw (-0.3ex,0) -- (0.3ex,0);
      \draw (0,-0.3ex) -- (0,0.3ex);
    \end{tikzpicture}%
  }}}%
}

\DeclareRobustCommand{\smallominus}{%
  \mathbin{\vcenter{\hbox{%
    \begin{tikzpicture}[baseline]
      \draw (0,0) circle (0.5ex);
      \draw (-0.3ex,0) -- (0.3ex,0);
    \end{tikzpicture}%
  }}}%
}

\newcommand{\question}[1]{{\fontfamily{ppl}\selectfont Q{\footnotesize{#1}}\xspace}}

\newcommand{\Textm}{\mathsf{T}}
\newcommand{\Imagem}{\mathsf{I}}
\newcommand{\Tadd}{\ensuremath{\Textm_{\smalloplus}}}
\newcommand{\Tsub}{\ensuremath{\Textm_{\smallominus}}}
\newcommand{\Iadd}{\ensuremath{\Imagem_{\smalloplus}}}
\newcommand{\Isub}{\ensuremath{\Imagem_{\smallominus  }}}
\newcommand{\Udig}{\ensuremath{\mU_{\mathrm{dig}}}}

\newcommand{\Dsp}{\texorpdfstring{\ensuremath{\mD}\xspace}{D}}
\newcommand{\Lsp}{\texorpdfstring{\ensuremath{\mL}\xspace}{L}}
\newcommand{\Pd}{\ensuremath{\Pi_{\Dsp}}}
\newcommand{\Pl}{\ensuremath{\Pi_{\Lsp}}}
\newcommand{\dD}{\ensuremath{\Delta\vh_{\Dsp}}}
\newcommand{\dL}{\ensuremath{\Delta\vh_{\Lsp}}}

\newcommand{\Wdig}{\ensuremath{\mW_{\mathrm{dig}}}}
\newcommand{\wdig}[1]{\ensuremath{\vw_{#1}}}
\newcommand{\wbar}{\ensuremath{\bar{\vw}}}
\newcommand{\Pdig}{\ensuremath{\Pi_{\Udig}}}
\newcommand{\Ogrp}[1]{\ensuremath{\mathrm{O}(#1)}}

\title{Beyond the Readout: A Latent Numerical Frame Shared Across Modality and Operation}

\author{%
  Oriol Juan \\
  Universitat Autònoma de Barcelona \\
  OriolJuan.Sabater@autonoma.cat
  \And
  Alex Ferrando de las Morenas \\
  Apple \\
  a\_ferrando@apple.com
  \And
  Pau Rodríguez \\
  Apple \\
  pau.rodriguez@apple.com \\
  \And
  Xavier Suau \\
  Apple \\
  xsuaucuadros@apple.com
}

\usepackage{tcolorbox}
\definecolor{takeawaybg}{RGB}{243, 246, 251}   
\definecolor{takeawayframe}{RGB}{196, 209, 228} 
\definecolor{takeawaytitle}{RGB}{35, 65, 110}    

\newtcolorbox{takeawaybox}[1][]{
    colback=takeawaybg,
    colframe=takeawayframe,
    boxrule=0.6pt,
    arc=2.5pt,
    left=7pt,
    right=7pt,
    top=5pt,
    bottom=5pt,
    boxsep=0pt,
    before=\par\vspace{10pt}\noindent,
    after=\par\vspace{10pt},
    #1
}
\begin{document}

\maketitle

\begin{abstract}
When a multimodal model computes $17+35$ from text or $91-39$ from a rendered image, it produces the same answer, $52$. Does it use one internal concept of number for both, or separate machinery for each path? Similarity metrics such as RSA or CKA cannot settle this question. They give the same score to identical and to rotated directions, and two representations can look alike only because both prepare the same output digits. We therefore use causal interventions. For four conditions (text or image, addition or subtraction), we learn a small subspace of \gemma's residual stream in which swapping coordinates swaps the model's answer. The subspaces answer three questions.
First, do the conditions write numbers along the same directions? They do not. Reusing a subspace from text in an image on the same operation recovers only $53.2\%$ of its causal effect. A single rotation and rescaling maps one subspace onto the other and raises recovery to $91.4\%$, so the conditions share one geometry expressed in different coordinates.
Second, does this shared geometry only reflect the digit the model is about to print? To test this, we split each subspace into $9$ directions that the output layer reads to choose the next digit and $13$ directions that it cannot see. The $13$ hidden directions still carry the number, yet changing them barely alters the digit being written.
Third, what are these hidden directions for? A model must emit a multi-digit answer one digit at a time, so it can compute the full result early and keep it in reserve. When two numbers share their first digit, swapping the number in the hidden directions leaves the digit being written intact and redirects the next digit in $77.6\%$ of cases. The model thus forms an abstract numerical concept upstream and unfolds it across time: multimodal representations share meaning through relational geometry across distinct coordinate frames.
\end{abstract}

\section{Introduction}
\label{sec:introduction}

When a multimodal language model computes an arithmetic result, it can reach the
same number through fundamentally different inputs and operations. For example,
the quantity $52$ can be reached by evaluating $17+35$ or $91-39$, presented as
plain text or rendered as pixels in an image. Because all of these inputs arrive
at the same mathematical truth, do they converge onto a shared internal
representation of number?

The Platonic Representation Hypothesis \citep{HuhEtAl2024} suggests they should,
positing that representations across diverse inputs converge toward a shared
statistical model of reality. Yet when comparing different computational routes
to the same number, ``representational convergence'' can mean three fundamentally
distinct physical possibilities:
\textbf{\question{1}: Coordinate identity (shared directions).} Does the network place numbers along the exact same linear directions regardless of input modality or operation? If so, an intervention discovered along one route will transfer directly to another without modification.
\textbf{\question{2}: Relational isomorphism (shared geometry in different coordinates).} Do different routes preserve the same relative distances and displacements between numbers, but orient them along different coordinate axes? If so, direct transfer fails, but a rigid rotation and scaling can re-align them.
\textbf{\question{3}: Output interface artifact.} Do representations for the same result under different conditions resemble each other only because the model is preparing to emit the exact same output token? If so, similarity reflects the shared digit vocabulary rather than an abstract concept of number.

Separating these hypotheses requires moving beyond passive similarity metrics.
Representational Similarity Analysis (RSA) \citep{KriegeskorteEtAl2008},
Centered Kernel Alignment (CKA) \citep{kornblith2019similarity}, and
Canonical Correlation Analysis (CCA) \citep{Hotelling1936RelationsBT} compare
internal relationship structures and are invariant to orthogonal rotations by construction. Consequently, they yield
identical scores under both \question{1} and \question{2}, making them incapable
of determining whether coordinates are shared or rotated. Furthermore,
similarity metrics are inherently blind to \question{3}: any two tasks sharing
an identical output vocabulary exhibit correlated late-layer representations
simply from anticipating the output interface. Most fundamentally, representational
agreement does not imply shared causal function \citep{UsamaChang2026} since models
can exhibit matching geometry while using divergent causal mechanisms.
Claiming convergence across modalities requires testing representations under
counterfactual interventions.

We resolve \question{1}, \question{2}, and \question{3} inside a single
instruction-tuned multimodal model, where three structural properties make them
cleanly separable. First, text and visual inputs pass through a unified backbone
and share a single residual stream, so coordinate identity (\question{1}) is
therefore an empirical question about one physical vector space, avoiding the
confounds of aligning two independently trained networks. Second, arithmetic
provides ground-truth counterfactual relations by construction: we can test
whether the vector displacement transforming $52$ into $71$ under addition in
text is the identical change under subtraction in images (\question{2}). Third,
because numbers are tokenized digit by digit, the ten decimal digit unembeddings
form an explicit rank-$9$ output subspace. By projecting out this readout
interface, we can study the remaining latent numerical geometry in isolation (\question{3}).

\begin{figure}[tb]
    \centering
    \includegraphics[width=\textwidth]{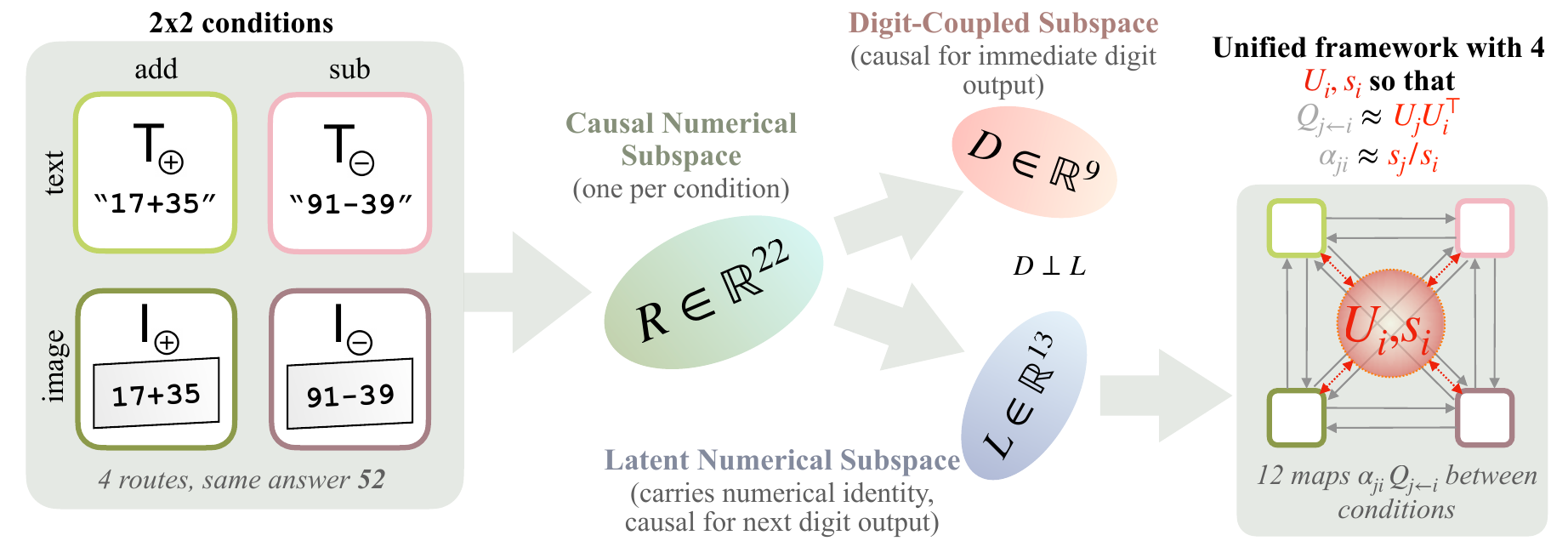}
    \caption{
        \textbf{Overview of the experimental framework.}
        Crossing input modality ($\Textm, \Imagem$) with operation
        ($\smalloplus, \smallominus$) yields four conditions reaching the same
        number. In each condition, we learn a $22$-dimensional causal subspace
        $\mR_i$ that governs generation (\cref{sec:setup}). While direct reuse
        fails across modalities, scaled rotations $\alpha\mQ$ restore causal
        control to $91.4\%$ (\cref{sec:transport}). Decomposing $\mR_i$ into a
        $9$-dimensional readout subspace $\Dsp$ and a $13$-dimensional latent
        subspace $\Lsp$ reveals that numerical geometry survives outside the
        digit readout and synchronizes into a single shared latent frame
        (\cref{sec:frame}). Finally, latent numerical identity in $\Lsp$ is
        scheduled across time, redirecting the token emitted at the subsequent
        autoregressive step (\cref{sec:autoregressive}).
    }
    \label{fig:overview}
\end{figure}

Crossing two modalities (text $\Textm$, image $\Imagem$) with two operations
(addition $\smalloplus$, subtraction $\smallominus$) produces four conditions
arriving at the same numerical result (\cref{fig:overview}). In each condition,
we use Distributed Alignment Search (\das) \citep{GeigerEtAl2023} to identify a
$22$-dimensional causal subspace $\mR_i$ that controls the generated answer.

Our contributions are the following three findings:
\textbf{(i) Coordinates differ, but numerical changes rotate (\cref{sec:transport}).}
The answer to \question{1} is no: direct reuse of a causal subspace across modalities recovers only $53.2\%$ of the intervention effect for same-operation pairs. However, \question{2} holds: corresponding numerical changes are related by scaled rotations ($\alpha\mQ$), elevating same-operation cross-modal recovery to $91.4\%$, generalizing across all twelve directed condition pairs, and composing across factors without refitting.
\textbf{(ii) Numerical geometry survives outside the digit readout in a shared frame (\cref{sec:frame}).}
Answering \question{3}, the shared structure is not an output artifact. Partitioning each causal subspace into a $9$-dimensional readout subspace $\Dsp$ and a $13$-dimensional latent subspace $\Lsp$ shows that ablating $\Dsp$ destroys immediate generation, yet $\Lsp$ preserves numerical geometry, retrieving unseen values at $75.5\%$ top-1 against $5\%$ chance. Orthogonal group synchronization shows the four latent spaces $\Lsp_i$ align into a single shared latent frame.
\textbf{(iii) Latent numerical identity steers generation one step later (\cref{sec:autoregressive}).}
The apparent paradox that $\Lsp$ encodes the number yet barely affects immediate output is resolved by timing. For pairs of numbers that share their first digit, swapping the number in $\Lsp$ leaves the current digit untouched ($0.965$ probability) but redirects the subsequent digit, matching the counterfactual value in $77.6\%$ of cases. The abstract number is formed upstream and read out sequentially across steps. A second multimodal model, \ministral, reproduces the full chain (\cref{app:ministral}).

\section{Related work}
\label{sec:background}

Our investigation builds on mechanistic studies of neural arithmetic, representation alignment, and causal mediation. Small networks trained on modular arithmetic learn periodic, Fourier-based solutions \citep{NandaEtAl2023Grokking,McCrackenEtAl2025}. In larger language models, arithmetic computation is distributed across attention heads and heuristics \citep{StolfoEtAl2023,QuirkeEtAl2024Arithmetic,ZhangEtAl2024Arithmetic,NikankinEtAl2025Arithmetic}, often embedding numbers via Fourier features \citep{ZhouEtAl2024}, cyclic base-10 circuits \citep{FeuchtEtAl2026,EngelsEtAl2025}, or helical manifolds \citep{KantamneniTegmark2025,FuEtAl2026}. To move beyond correlational probing, activation patching and causal mediation isolate hidden components responsible for model predictions \citep{VigEtAl2020,StolfoEtAl2023}. Distributed Alignment Search (\das) \citep{GeigerEtAl2023} formalizes this by learning linear subspaces aligned with high-level variables. Recently, studies have documented that internal representations can compute abstract information prior to token emission, formalizing this as latent planning \citep{HannaAmeisen2026,GurneeEtAl2026Workspace,LindseyEtAl2025}. Finally, the Platonic Representation Hypothesis \citep{HuhEtAl2024} posits that representations across diverse modalities converge toward a shared statistical model of reality, though recent work cautions that representational similarity alone does not imply shared causal mechanism \citep{UsamaChang2026}.

\section{Experimental setup and causal subspaces}
\label{sec:setup}

To evaluate how numbers are represented across modalities and operations, the
experimental framework must eliminate confounding factors: modalities must
share a single vector space, conditions must evaluate identical mathematical
quantities, and representations must exert causal control over generation.

\textbf{Model and experimental conditions.}
We evaluate the instruction-tuned \gemma \citep{GemmaTeam2026}, a multimodal
transformer with $48$ layers and residual stream dimensionality
$d_{\mathrm{model}}=3840$. Because text and image inputs are projected into a
single residual stream, representations across modalities inhabit the same
physical vector space. We disable thinking mode and prompt the model to emit
only the numerical result, ensuring no intermediate chain-of-thought tokens enter
the residual stream.

We establish four conditions crossing input modality with arithmetic operation:
Text Addition ($\Tadd$), Text Subtraction ($\Tsub$), Image Addition ($\Iadd$),
and Image Subtraction ($\Isub$), where $\Textm$ and $\Imagem$ denote text and
image modalities, and $\smalloplus$ and $\smallominus$ denote the operations.
Operands $a, b$ are integers sampled from $[0, 99]$ with valid domains
$a + b \le 99$ for addition and $a \ge b$ for subtraction, yielding exactly
$5050$ expressions per operation. For image inputs, the identical arithmetic
expressions are rendered as centered black text on a clean white canvas. To
ensure representational comparisons reflect input format rather than task
competence, evaluations are restricted to expressions the model answers
correctly across both modalities (\cref{app:setup}).

\textbf{Extracting causal subspaces via Distributed Alignment Search.}
Let $\vh_{\ell,t}(x) \in \R^{d_{\mathrm{model}}}$ denote the residual stream
activation at layer $\ell$ and prompt position $t$. We extract representations at
layer $\ell=43$ at the final token position $t$ immediately preceding generation.
Linear probing, activation patching, and counterfactual interventions identify a late-layer region with strong numerical representations and causal influence. We use layer 43 as a common reference across conditions, and the final pre-generation position provides a semantically comparable intervention site across modalities
(\cref{app:reference-config,app:visual}).

In each condition, we use \das \citep{GeigerEtAl2023} to learn an orthonormal
basis $\mR \in \R^{d_{\mathrm{model}} \times k}$ (with $\mR^\top\mR = \mI_k$)
spanning a $k$-dimensional causal subspace. With the model weights completely
frozen, $\mR$ is optimized by gradient descent so that swapping coordinates
inside the subspace replaces the base answer with the donor's answer. Formally,
given a base activation $\vh_b$ and a donor activation $\vh_d$ carrying a
counterfactual result, the interchange intervention is:
\begin{equation}
\label{eq:das}
\vh_b' = \vh_b + \mR\left(\mR^\top \vh_d - \mR^\top \vh_b\right),
\end{equation}
which modifies only coordinates spanned by $\mR$ while leaving the orthogonal
complement $(\mI - \mR\mR^\top)\vh_b$ untouched.

We quantify causal efficacy using the autoregressive interchange intervention
accuracy ($\iiaar$): the fraction of interventions after which the model freely
generates the complete multi-digit target integer. To select the subspace
dimension $k$, we evaluate causal retention relative to  $k=32$ baseline, which achieved the highest $\iiaar$ in a preliminary dimension sweep. We select $k=22$ as the smallest tested dimension retaining at least 90\% of its autoregressive IIA across all four conditions (\cref{app:das-training}).

\textbf{Why causal subspaces instead of Fourier planes?}
Prior studies emphasize periodic Fourier features as natural coordinates for
arithmetic in language models \citep{ZhouEtAl2024,KantamneniTegmark2025,NikankinEtAl2025Arithmetic}.
In \gemma, however, pure Fourier features are causally insufficient: the $11$
most Fourier-aligned directions in the causal subspace recover only $23\%$ of the
full autoregressive intervention effect, compared with $84\%$ in \llama
(\cref{app:fourier}). Because the causal numerical state in modern multimodal
models is substantially more distributed, we analyze the complete learned
causal subspaces.

\begin{takeawaybox}
\textbf{Takeaway 1 (Causal Subspaces):}
 Intervening on Fourier-aligned directions alone fails to recover most of the causal effect ($23\%$ recovery). The full 22-dimensional causal subspaces $\mR_i$ achieve $ 69.3\%-86.2\%$ autoregressive accuracy, identifying the directions mediating arithmetic answers.
\end{takeawaybox}

\section{Numerical change transports across modality and operation}
\label{sec:transport}

If the four conditions encoded numbers along identical residual-stream directions,
a causal subspace learned in one condition would transfer directly to another.
Evaluating direct subspace reuse therefore provides a decisive empirical test of
coordinate identity (\question{1}). When direct reuse fails, the mathematical
structure of that failure reveals what coordinate transformation is missing
(\question{2}).

\textbf{Measuring direct reuse across conditions.}
Transporting a causal intervention tests whether steering an answer in text operates through the same physical residual-stream directions as steering an answer in vision.
Let condition $A$ denote the source where causal subspace $\mR_A$ was trained,
and condition $B$ denote the destination where it is evaluated. Direct reuse
inserts $\mR_A$ into destination inputs using activations drawn entirely from $B$,
ensuring no source activations enter destination computation. Because baseline
intervention difficulty varies across conditions, raw intervention scores would
conflate alignment with intrinsic destination difficulty. We therefore measure
the normalized recovery $\tsc{A}{B}$:
\begin{equation}
\label{eq:normalized-transfer}
\tsc{A}{B}
=
\frac{\iiaar(\mR_A\rightarrow B) - \iiaar(\Rrand\rightarrow B)}
     {\iiaar(\mR_B\rightarrow B) - \iiaar(\Rrand\rightarrow B)},
\end{equation}
where $\iiaar(\mR_A\rightarrow B)$ denotes the intervention accuracy of the reused
subspace, and $\Rrand$ is a random $22$-dimensional subspace sampled from the
principal component span of destination activations. A score of $\tau=1$
indicates full recovery of the native causal effect, while $\tau=0$ denotes
performance no better than random chance. Because $\tau$ is normalized by the
destination's native subspace, values can exceed $1$ if the foreign subspace
happens to steer the destination more effectively. All reported values are
averaged over $3\times3$ \das training seed pairs.

\begin{figure}[htb]
    \centering
    \includegraphics[width=\linewidth]{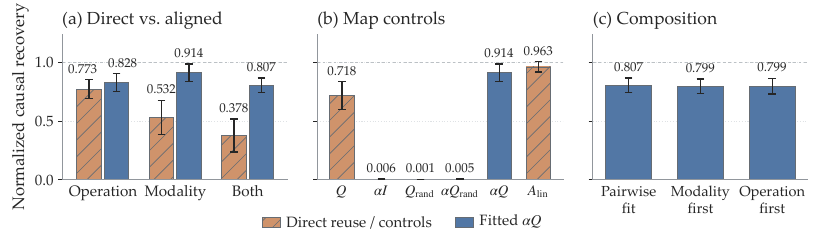}
    \caption{
        \textbf{Causal transfer across conditions.}
        \textbf{(a)} Direct versus aligned reuse (fitting $\alpha \mQ$) for changes in operation, modality, or both.
        \textbf{(b)} Cross-modal transformation controls. A map of the form $\alpha \mQ$ aligns subspaces nearly as well as an unconstrained map. Rotation alone improves recovery, and the fitted scale $\alpha$ provides a further gain.
        \textbf{(c)} Pairwise and composed transports. Bars average four direction means. Scores are destination-normalized autoregressive recovery on held-out expressions.}
    \label{fig:transport}
\end{figure}

\textbf{Direct reuse is incomplete across modalities.}
Within a single modality, direct subspace reuse across operations is remarkably
strong: $\tsc{A}{B}$ averages $77.3\%$ (\cref{fig:transport}a). However, across
modalities direct reuse drops substantially: the eight cross-modal condition pairs
range from $23.4\%$ to $69.2\%$, averaging $45.5\%$. Furthermore, cross-modal
transfer is directional: subspaces trained on images recover $57.7\%$ of the causal
effect in text, whereas text-trained subspaces recover only $33.4\%$ in images.
Restricting to same-operation cross-modal pairs ($\Tadd \leftrightarrow \Iadd$
and $\Tsub \leftrightarrow \Isub$), direct reuse averages $53.2\%$
(\cref{fig:transport}a; \cref{app:direct-transfer}). 

These findings resolve \question{1} in the negative: \emph{the four conditions do
not share identical coordinate directions in the residual stream.}

\textbf{Scaled rotations restore causal control.}
If coordinate directions differ, can they be aligned by a geometric mapping?
Instead of aligning static activation vectors by replacement, which confounds representation
with baseline shift, we align the quantity modified by the intervention: the
counterfactual displacement vector. For a base-donor pair in source condition
$A$, the displacement in subspace coordinates is
$\Delta \vz_A = \mR_A^\top(\vh_{d,A} - \vh_{b,A})$. Given the corresponding
displacement $\Delta \vz_B$ in destination condition $B$, we fit a scaled
rotation $(\alpha^\star, \mQ^\star)$ via constrained least squares:
\begin{equation}
\label{eq:scaled-procrustes}
(\alpha^\star, \mQ^\star)
=
\argmin_{\alpha,\;\mQ}
\sum_i \left\| \Delta \vz_B^{(i)} - \alpha\mQ\,\Delta \vz_A^{(i)} \right\|_2^2
\quad \text{s.t.} \quad \mQ^\top\mQ = \mI_k,
\end{equation}
where $\mQ \in \Ogrp{k}$ is an orthogonal matrix that rotates displacements
without shearing, and $\alpha > 0$ is a single scalar that rescales overall
magnitude uniformly across all directions. The transported intervention is then
applied to the destination as
$\vh'_{b,B} = \vh_{b,B} + \mR_B\,\alpha^\star\mQ^\star \Delta \vz_A$, evaluated
strictly on held-out numerical expressions.

An orthogonal rotation alone substantially improves cross-modal transfer, achieving
$0.718$ normalized recovery compared with $0.532$ for direct reuse
(\cref{fig:transport}b; \cref{app:alignment-targets}). Adding a single uniform
scale $\alpha$ further increases mean recovery by $0.196$ to $0.914$ across the
four same-operation cross-modal directions (\cref{fig:transport}a,b). The transported
displacements achieve a cosine similarity of $0.902$ for addition and $0.915$ for
subtraction with true destination displacements. Control experiments show that
rotation alone is effective, while scaling provides a substantial additional gain:
scaling without rotation yields only $0.006$ recovery, and a random rotation
fails with or without scaling (\cref{fig:transport}b). Finally, replacing the
scaled rotation with an unrestricted linear map yields only a minor gain from
$0.914$ to $0.963$ (\cref{app:transport-ablations}).

\textbf{Visual representations are geometrically compressed.}
The learned scale parameters $\alpha$ reveal a clear physical regularity: their
magnitude is dictated by the direction of modality transfer instead of the
arithmetic operation (\cref{tab:app-transport-ablation,app:transport-ablations}). Displacements are tightly estimated
at $\alpha = 0.639 \pm 0.006$ for $\Tadd \to \Iadd$ and $0.658 \pm 0.011$ for
$\Tsub \to \Isub$ when mapping from text into images. In reverse, mapping from
images into text yields scales of $1.283 \pm 0.019$ and $1.271 \pm 0.021$.
Numerical displacements in the visual modality are systematically smaller in
magnitude than their textual counterparts ($\alpha \approx 0.65$). This magnitude
difference may contribute to the asymmetry observed under direct reuse, while the
fitted scales alone do not establish its architectural cause.

\textbf{Compositionality across modality and operation.}
We next evaluate condition pairs where the arithmetic operation changes.
Within a single modality ($\Tadd \leftrightarrow \Tsub$ or $\Iadd \leftrightarrow \Isub$),
conditions are matched by the numerical transition they implement ($a \to b$).
Here, direct subspace reuse already achieves mean $0.773$ normalized recovery ($\tau$),
demonstrating that switching operations preserves substantial coordinate compatibility
without requiring alignment. Fitting explicit scaled rotations across operations
reaches mean $0.828$ recovery (\cref{fig:transport}a).
Crucially, these geometric maps compose across experimental factors.
For condition pairs that change both modality and operation simultaneously (e.g., $\Tadd \to \Isub$),
directly fitted maps recover mean $0.807$.
Without fitting this diagonal relation directly, chaining a modality rotation with an operation
rotation ($\Tadd \to \Iadd \to \Isub$ or $\Tadd \to \Tsub \to \Isub$) achieves $0.799$ and $0.799$
recovery (\cref{fig:transport}c), yielding displacement predictions with mean cosine
similarities of $0.968$ and $0.969$ to the directly fitted maps
(\cref{tab:app-composed-paths,app:cross-operation}).

These results resolve \question{2} affirmatively: \emph{numerical changes share an
isomorphic geometry related by scaled rotations.} However, because these causal
subspaces are measured at the final prompt token before generation, an essential
question remains: does this shared geometry represent the abstract number, or
merely the digit token the model is preparing to emit? This brings us to
\question{3}.

\begin{takeawaybox}
\textbf{Takeaway 2 (Scaled Rotations Align Modalities):}
Direct subspace reuse across modalities is incomplete ($53.2\%$ recovery), refuting coordinate identity (\question{1}). Scaled rotations $\alpha\mQ$ restore causal steering to $91.4\%$ (\question{2}), with $\alpha \approx 0.65$ capturing visual representational compression.
\end{takeawaybox}

\section{Beyond readout: a shared latent frame}
\label{sec:readout}
\label{sec:frame}

   \question{3} asks whether the alignment of \cref{sec:transport} reflects the output interface.
   All four conditions are read at the token preceding generation, so all of them are about to
   emit one decimal digit. If the shared geometry lived only in the directions that separate
   these digits, it would reflect tokenization and would say nothing about number. We therefore
   measure how strongly the causal subspaces couple to the digit readout, remove that coupling,
   and test whether numerical geometry survives.

\textbf{Causal subspaces are strongly coupled to digit unembeddings.}
Let $\Wdig \in \R^{d_{\mathrm{model}} \times 10}$ collect the unembedding vectors
corresponding to the decimal digits $0$ through $9$. To isolate directions that
differentiate digits from each other, leaving out the shift-invariant component
common to all digit logits, we subtract their mean vector $\wbar$ and orthonormalize
the result. Because these centered vectors sum to zero, they span at most $9$
dimensions, yielding the rank-$9$ digit readout basis
$\Udig = \operatorname{orth}(\{ \wdig{d} - \wbar \}_{d=0}^{9})$.

The learned causal subspaces $\mR_i$ exhibit substantial geometric alignment with
$\Udig$ (\eg exceed random residual-stream baselines by a factor of $41.65$
at layer $43$ in text addition). Although the projection onto $\Udig$ accounts for
$9.8\%$ to $11.0\%$ of the total squared Frobenius norm of $\mR_i$, that modest
fraction carries virtually all immediate causal power over the next token.
Projecting $\Udig$ out of the causal subspace causes self-intervention accuracy
$\iiaar$ to collapse: performance drops from $0.826$, $0.865$, $0.747$, and
$0.695$ to $0.008$, $0.000$, $0.005$, and $0.003$ for $\Tadd$, $\Tsub$, $\Iadd$,
and $\Isub$ respectively. In contrast, ablating a random rank-$9$ subspace
preserves substantial causal efficacy ($0.634$, $0.791$, $0.339$, and $0.560$;
\cref{fig:app-readout}b). A non-arithmetic control task requiring the model only
to echo a single digit (\texttt{Repeat this number: 7 =}) confirms that this
coupling is functionally specialized for digit production (\cref{app:digit-readout}).

\textbf{An exact decomposition: Readout subspace $\Dsp$ and Latent subspace $\Lsp$.}
To isolate what exists outside the output interface, projecting the digit span out
of the entire residual stream via $\Pdig^\perp = \mI - \Udig\Udig^\top$ leaves
cross-condition numerical geometry intact in the DAS representation
(\cref{app:readout-free}). However, whole-stream projection is a blunt instrument:
it alters coordinates across the entire $3840$-dimensional space.

We therefore construct an exact internal decomposition within the learned
$22$-dimensional causal subspace $\mR_i$. We rotate the basis vectors of $\mR_i$
to order them by their alignment with the digit readout span $\Udig$, and split
the space at rank $9$ (the rank of $\Udig$). This partitions $\mR_i$ into two
orthogonal subspaces:
a $9$-dimensional \textbf{readout subspace} $\Dsp \subset \operatorname{col}(\mR_i)$ containing all coupling between the causal subspace and the digit unembedding, and an orthogonal $13$-dimensional \textbf{latent subspace} $\Lsp \subset \operatorname{col}(\mR_i)$ in the exact null space of the digit unembedding (overlap $O(10^{-7})$; \cref{app:readout-free}). The dimension of $\Lsp$ is fixed by the $k=22$ causal subspace (fixed for the main analyses) minus the rank-$9$ unembedding span, leaving $13$ dimensions that encode numerical values across the full range $[0, 99]$ without coupling to immediate token logits.
Defining $\Pd$ and $\Pl$ as the orthogonal projectors onto $\Dsp$ and $\Lsp$
(they keep only the component of a vector lying in each subspace), these projectors
satisfy $\Pd + \Pl = \mR_i\mR_i^\top$, so the split of $\mR_i$ is exact. Note that
$\Dsp$ is a $9$-dimensional subspace of $\mR_i$ chosen to align with $\Udig$. It is
close to $\Udig$ but not identical to it, because $\Udig$ lives in the full residual
stream while $\Dsp$ lives inside the causal subspace.

\begin{figure}[htb]
    \centering
    \begin{minipage}[t]{0.99\textwidth}\vspace{0pt}%
        \includegraphics[width=\linewidth, clip]%
            {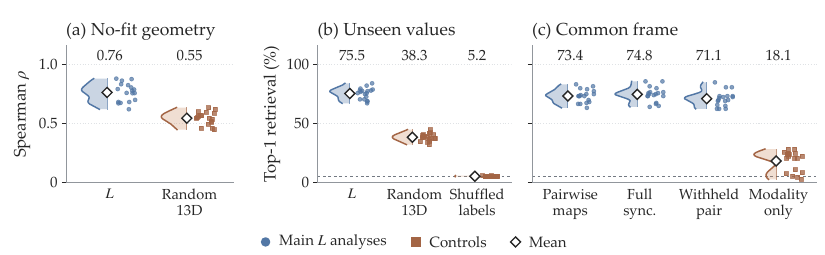}%
    \end{minipage}
    \caption{\textbf{Numerical geometry beyond readout supports a common frame.}
    \textbf{(a)} No-fit RSA in $\Lsp$ ($\rho = 0.765$) versus random $13$-dimensional readout-orthogonal subspaces ($\rho = 0.545$).
    \textbf{(b)} Unseen-value retrieval after alignment ($75.5\%$), with random-space ($38.3\%$) and shuffled-label ($5.2\%$) controls.
    \textbf{(c)} Pairwise maps ($73.4\%$) versus full synchronization ($74.8\%$), withheld-pair reconstruction ($71.1\%$), and modality-only synchronization ($18.1\%$). Points average repeats within each relation and value split; half-violins show distributions, and diamonds and numbers show means. The dashed line marks the $5\%$ chance baseline among $20$ candidate values.}
    \label{fig:L-geometry}
\end{figure}

\textbf{Numerical geometry survives inside the latent subspace $\Lsp$.}
Having cleanly isolated the $13$-dimensional latent subspace $\Lsp$ from the
output readout, we test whether it retains structured numerical information.
We project activations at layer $43$ into each condition's latent subspace
$\Lsp_i$ and evaluate cross-condition alignment on values held out during alignment fitting.

The latent subspaces exhibit rich numerical geometry across all twelve directed relations: mean held-out transition cosine reaches $0.807$ (\cref{tab:app-readout-free}), and nearest-neighbor retrieval identifies the exact unseen value among $20$ candidates at $75.5\%$ top-1 and $98.8\%$ top-5 accuracy (\cref{fig:L-geometry}b; $5\%$ random baseline). In contrast, dimension-matched random $13$-dimensional subspaces in the readout null space achieve $0.558$ transition cosine and $38.3\%$ top-1 retrieval, while shuffling numerical labels collapses retrieval to chance ($5.2\%$). A no-fit RSA check likewise yields rank correlation $\rho = 0.765$ across conditions (permutation $p = 1/1001$) versus $0.545$ for random subspaces (\cref{fig:L-geometry}a).

These findings support a negative answer to \question{3}:
\emph{the shared numerical representation is not an artifact of the shared digit
interface.} The relational geometry of arithmetic survives within an internal
subspace that the current digit readout cannot see.

\textbf{One common latent frame across all conditions.}
The twelve pairwise relations are fitted independently. Could they describe twelve
unrelated mappings that happen to align pairwise? A stronger hypothesis predicts
that every condition's latent subspace $\Lsp_i$ reflects one unified underlying
frame: each condition possesses an intrinsic orientation $\mU_i \in \Ogrp{13}$
and scale $s_i > 0$ such that pairwise rotations factorize as
$\mQ_{j\leftarrow i} \approx \mU_j\mU_i^\top$ and scales factorize as
$\alpha_{ji} \approx s_j / s_i$.

We estimate the four frames jointly using orthogonal group synchronization
\citep{GaoZhang2023}, minimizing total disagreement across all twelve pairwise
maps simultaneously:
\begin{equation}
\label{eq:sync-objective}
\{\mU_i^\star\}
=
\argmin_{\{\mU_i\}}
\sum_{(i,j)} \xi_{ij}
\left\| \mQ_{j\leftarrow i} - \mU_j\mU_i^\top \right\|_F^2 ,
\end{equation}
with edge confidences $\xi_{ij}$ and scales $s_i$ solved by weighted least
squares on $\log\alpha_{ji} \approx \log s_j - \log s_i$. If pairwise linear
relationships were exact and cycle-consistent, a shared global frame would follow algebraically.
Real activations, however, contain noise and task-specific variance. Chaining
independently fitted maps around a closed loop typically accumulates drift
instead of returning to the identity. Constraining all twelve directed maps to
factor through four node frames $\mU_i$ imposes a fourfold parameter reduction
(from $948$ to $237$ free parameters), testing whether the underlying representations
are governed by a globally consistent low-noise geometry rather than isolated
pairwise compromises. To map absolute numerical coordinates in addition to relative
displacements, we estimate one centroid offset $\vo_i$ per condition from training
values:
\begin{equation}
\label{eq:frame-absolute}
\vu(y) = s_i^{-1}\mU_i^\top\left( \vz_i^{\Lsp}(y) - \vo_i \right) ,
\qquad
\vz_j^{\Lsp}(y) = s_j\mU_j\vu(y) + \vo_j .
\end{equation}
Taking differences cancels the offset vectors, recovering the displacement
relation $\Delta \vu = s_i^{-1}\mU_i^\top \Delta \vz_i^{\Lsp}$. The synchronization
in \cref{eq:sync-objective} therefore depends strictly on numerical displacements,
and offsets are incorporated afterwards to locate absolute coordinates (\cref{app:sync-protocol}).

On unseen numerical values, the four synchronized frames achieve $0.791$
transition cosine and $74.8\%$ top-1 retrieval, comparable to independently fitted
maps ($0.801$ and $73.4\%$; \cref{fig:L-geometry}c). Preserving held-out
retrieval while eliminating $711$ degrees of freedom supports an underlying,
globally coherent relational frame. When one complete unordered condition pair is
withheld prior to synchronization, reconstructing it from the remaining ten
relations reaches $0.762$ cosine, $71.1\%$ top-1, and $98.3\%$ top-5 retrieval
(\cref{fig:L-geometry}c; \cref{tab:app-L-sync}). Furthermore, forcing operations to share frames within
modalities reduces cosine to $0.249$ and retrieval to $18.1\%$ (\cref{fig:L-geometry}c), supporting
condition-dependent latent frames.

Predicting coordinates for an unseen condition by adding modality and operation
shifts achieves $0.629$ cosine and $58.9\%$ top-1 retrieval on unseen values,
against $-0.109$ cosine and $3.8\%$ top-1 under shuffled controls (\cref{app:heldout-condition}).
An orthogonal successor operator $G_{+1}$ fitted only in $\Tadd$ on the
full $22$-dimensional causal spaces transfers without refitting, identifying
$y+1$ centroids in $65.0\%$ to $70.0\%$ of held-out cases (\cref{app:successor-equivariance}). Finally, synchronizing the
full $22$-dimensional causal subspaces yields $\iiaar = 0.636$ (versus $0.639$
for pairwise maps), preserving similar causal efficacy under global constraints
(\cref{app:full-das-sync}).

These findings establish the strong form of \question{2}: \emph{the latent numerical
representations across all four conditions express one shared relational frame.}

\begin{takeawaybox}
\textbf{Takeaway 3 (A Shared Latent Frame Beyond Readout):}
Abstract numerical geometry lives in the $13$d latent subspace $\Lsp$, orthogonal to digit unembeddings (\question{3}). The four conditions synchronize into a single common latent frame ($74.8\%$ top-1 retrieval of values held out from alignment fitting).
\end{takeawaybox}

\section{Latent numerical identity steers subsequent generation}
\label{sec:autoregressive}

The findings of the preceding sections create an apparent paradox: the latent
subspace $\Lsp$ is organized by numerical identity, aligns across modalities and
operations, and synchronizes into a single common frame. Yet intervening on
$\Lsp$ alone recovers the complete donor answer in only $3.2\%$ to $6.4\%$
of cases, depending on the condition. How can a representation reliably encode the mathematical answer
while exerting little causal influence on immediate output?

The architecture of an autoregressive model requires emitting tokens sequentially, but solving arithmetic requires computing the entire multi-digit answer upstream. We test whether the network resolves this tension through a temporal division of labor: holding the full numerical identity in $\Lsp$ while $\Dsp$ governs only the immediate token. We design a mismatched intervention testing this temporal division: $\Dsp$  controls the token written \emph{now}, while $\Lsp$  steers the token written \emph{next}.

\textbf{Constructing a mismatched intervention state.}
The orthogonal decomposition $\mR_i\mR_i^\top = \Pd + \Pl$ allows us to
synthesize hybrid residual activations whose two components originate from
different numerical answers. Let $\vh_b$ be a base activation evaluating result
$b$, and let $\vh_y$ and $\vh_{\tilde y}$ be donor activations evaluating $y$
and $\tilde y$. We decompose donor edits into readout displacements
$\dD^{b\to y} = \Pd(\vh_y - \vh_b)$ and latent displacements
$\dL^{b\to \tilde y} = \Pl(\vh_{\tilde y} - \vh_b)$. When both halves originate
from the same donor ($y = \tilde y$), adding both to $\vh_b$ reconstructs the
standard intervention of \cref{eq:das}:
$\vh^{\mathrm{match}} = \vh_b + \dD^{b\to y} + \dL^{b\to y}$.

In contrast, we construct a \textbf{mismatched state} by pairing the readout
displacement for $y$ with the latent displacement for $\tilde y$:
\begin{equation}
\label{eq:mismatched-state}
\vh^{\mathrm{mis}} = \vh_b + \dD^{b\to y} + \dL^{b\to \tilde y}.
\end{equation}
In this state, $\Dsp$ points to $y$, while $\Lsp$ points to $\tilde y$. We
select pairs $y$ and $\tilde y$ that share their first digit and differ only in
their second digit (e.g., $y=52$ and $\tilde y=57$). Both components therefore
agree on the tens digit due immediately, while disagreeing on the units digit due
at the next position.

\begin{figure}[htb]
    \centering
    \includegraphics[width=0.98\textwidth]{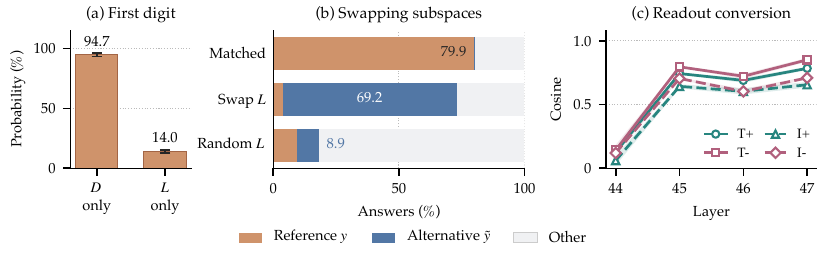}
    \caption{
        \textbf{The latent subspace $\Lsp$ sets the digit written next.}
        Each trial edits the activation of a \emph{base} prompt (\eg $17{+}14=31$) at layer $43$ of the last prompt position $t_0$, using donors with answers $y$ and $\tilde y$ that share their first digit and differ in the second (\eg $52$ and $57$).
        \textbf{(a) Current digit.}
        The first digit of $y$ differs from that of the base. Editing $\Dsp$ alone with the donor $y$ makes the model write the first digit of $y$ with probability $94.7\%$, editing $\Lsp$ alone only $14.0\%$.
        \textbf{(b) Full answer.}
        \emph{Matched} takes $\Dsp$ and $\Lsp$ from $y$ and yields $y$ in $79.9\%$ of cases, the ceiling of a $22$-dimensional edit.
        \emph{Swap $\Lsp$} takes $\Dsp$ from $y$ and $\Lsp$ from $\tilde y$ (\cref{eq:mismatched-state}), and yields $\tilde y$ in $69.2\%$.
        \emph{Random $\Lsp$} takes $\Lsp$ from an unrelated value and yields $\tilde y$ in only $8.9\%$.
        \textbf{(c) Delayed readout.}
        At the next step $t_1$, the cosine between the swap's effect and the natural change in readout directions is near zero at layer $44$ and jumps at layer $45$, once the edited KV-cache is read.
    }
    \label{fig:latent-to-readout}
\end{figure}

\textbf{Mismatched latent identity redirects the subsequent token.}
Across four conditions and three seeds, the matched state $\vh^{\mathrm{match}}$
produces target $y$ with mean accuracy $\iiaar = 0.799$ (\cref{fig:latent-to-readout}b). In contrast, the
mismatched state $\vh^{\mathrm{mis}}$ rarely produces $y$ ($\iiaar = 0.037$) and
produces $\tilde y$, supplied through $\Lsp$, at $0.692$ accuracy (versus $0.089$
for random controls; \cref{fig:latent-to-readout}b). The first-digit probability remains largely unchanged:
$0.963$ matched versus $0.965$ mismatched (\cref{app:latent-swap}).
A separate different-first-digit evaluation shows stronger immediate control from
$\Dsp$ alone ($94.7\%$) than from $\Lsp$ ($14.0\%$; \cref{fig:latent-to-readout}a).
With pairs differing in both digits (\eg $y=52$, $\tilde y=67$), the swap yields the hybrid (\eg $57$) in $51.6\%$ of cases, against $7.5\%$ for $\Dsp$ alone and $74.5\%$ for matched edits reproducing $y$ (\cref{tab:app-both-digits}). Random $\Lsp$ also inserts its own units digit (\cref{app:latent-swap}).

\textbf{Tracing the delayed identity through the KV-cache.}
We trace autoregressive generation across step $t_1$ after intervening at layer
$43$ of position $t_0$. Preserving the intervened KV-cache, we feed the model the
identical, valid first digit, and observe token logits at $t_1$ with no further
intervention ($1536$ separately sampled triplets; \cref{app:cache-propagation}). In the matched
state, the model continues with the units digit of $y$ ($P(y_2) = 0.824$ versus
$P(\tilde y_2) = 0.010$). In the mismatched state, the probabilities invert:
$P(y_2) = 0.040$ while $P(\tilde y_2) = 0.737$, and the units digit of
$\tilde y$ becomes the greedy top-1 prediction in $77.6\%$ of cases. Supplying $\Dsp$ alone without $\Lsp$ leaves the second-digit probabilities 
unresolved ($P(y_2) = 0.132$, $P(\tilde y_2) = 0.087$), showing that $\Dsp$ 
alone cannot steer subsequent digits. Likewise, inserting an $\Lsp$  
from an unrelated donor sharing the first digit leaves $P(\tilde y_2)$ at 
only $0.020$.

\textbf{Latent identity converts into readout directions via intervening layers.} Because $\Lsp$ is orthogonal to the digit readout $\mU_D$ when inserted after layer $43$ of step $t_0$, steering the second digit at step $t_1$ requires subsequent layers to rotate this latent edit into $\mU_D$. To track this conversion at each layer $\ell$ of step $t_1$, we compare how swapping $y \to \tilde y$ in $\Lsp$ shifts the readout coordinates, $\mU_D^\top(\vh_{\ell,t_1}^{\mathrm{mis}} - \vh_{\ell,t_1}^{\mathrm{match}})$, against the readout difference between unintervened runs evaluating $\tilde y$ and $y$, $\mU_D^\top(\vh_{\ell,t_1}^{\mathrm{nat}}(\tilde y) - \vh_{\ell,t_1}^{\mathrm{nat}}(y))$ (\cref{app:cache-propagation}). Their cosine similarity $\gamma_\ell$ starts near zero at layer $44$ ($\gamma_{44} \approx 0.113$), jumps sharply at layer $45$ ($\gamma_{45} \approx 0.722$) once attention reads the modified $t_0$ KV-cache, and reaches a median of $0.852$ in the final layer (\cref{fig:latent-to-readout}c). Moreover, projecting a non-arithmetic copy task (\texttt{Repeat this number: 43 =}) into $\Lsp$ reveals the same latent geometry even without computation (\cref{app:repeat-number}). We therefore interpret $\Lsp$ as an upstream state that holds an abstract multi-digit result and converts it into readout directions sequentially across steps.

\begin{takeawaybox}
\textbf{Takeaway 4 (Temporal Scheduling of Latent Identity):} The latent subspace $\Lsp$ has little direct influence on the current digit but steers the next digit ($77.6\%$ top-1 redirection after teacher-forcing the shared first digit). $\Lsp$ thus acts as a latent memory that buffers the full multi-digit result upstream and converts it into readout directions across autoregressive steps.

\end{takeawaybox}

\section{Discussion and Conclusion}
\label{sec:discussion}

\textbf{Representations can share causal meaning without sharing coordinate axes.}
When a model adds or subtracts numbers in text or images, it computes
an internal representation that behaves the same way across all four tasks.
Yet this shared representation does not occupy identical coordinates in the
residual stream. Directly copying the subspace across conditions
recovers only $45.5\%$ of the causal effect on average (\question{1} negative).
Instead, the spaces share a geometry: a single scaled rotation maps the axes of
one condition onto another and recovers $91.4\%$ of the causal effect, and all
four conditions synchronize into a single relational frame (\question{2}
affirmative). This distinction matters for how we study foundation models.
Geometry alone cannot determine whether two internal states do the same work, we need
interventions to verify their causal role \citep{UsamaChang2026}.

\textbf{Computation upstream, readout over time.}
This relational convergence lives outside the vocabulary. The shared
$13$-dimensional subspace $\Lsp$ is orthogonal to the digit unembeddings,
ruling out an output interface artifact (\question{3} refuted). While $\Dsp$
drives the immediate token, $\Lsp$ holds the full numerical identity in memory
and steers the subsequent token through the attention cache. The model decouples
upstream computation from token generation, solving the entire arithmetic
problem first and reading the answer out sequentially across autoregressive
steps.

\textbf{Scope and open questions.}
This mechanism extends beyond \gemma. A replication in \ministral
(\cref{app:ministral}) reproduces the readout-orthogonal geometry, global
synchronization, and delayed causal influence, showing that the division of
labor extends across model families. Our experiments focus on addition and
subtraction of integers in $[0, 99]$. Multiplication forms distributed,
depth-dependent representations across intermediate layers
(\cref{app:multiplication}). Finally, while we show that the delayed signal
reaches the output via the key-value cache, identifying the specific attention
heads that perform this readout remains an open direction.

\subsection*{AI use statement}
Large language models were used as writing and coding assistants, for editing prose and \LaTeX{}, drafting analysis and plotting scripts, and literature search. The authors designed the experiments, ran and verified all results, and take full responsibility for the content. The language models studied in this work (\gemma, \llama, and \ministral) are the objects of the experiments.

\subsection*{Ethics statement}
This work analyzes the internal activations of publicly available language models on synthetic arithmetic prompts (integers in $[0,99]$, presented as text and as rendered images). It involves no human subjects, personal data, or crowd-sourcing, and we foresee no direct societal risk. The interpretability methods described may help audit model behavior.

\subsection*{Reproducibility statement}
All models are public checkpoints (\cref{app:setup,app:ministral}). Prompts are generated programmatically from the integer range $[0,99]$ following the procedure in \cref{app:setup}. Layers, token positions, subspace dimensions, training hyperparameters, seeds, and controls are reported in the appendices. Code will be released soon.



\bibliography{references}
\bibliographystyle{iclr2027_conference}

\clearpage

\pretocmd{\section}{\FloatBarrier}{}{}
\appendix

\crefalias{section}{appendix}
\crefalias{subsection}{appendix}
\crefalias{subsubsection}{appendix}

\section{Experimental setup}
\label{app:setup}

\subsection{Prompts, datasets, and activation extraction}
\label{app:experimental-setup}

\textbf{Prompts.}
Thinking mode is disabled throughout and prompts request only the numerical
answer. For textual inputs, the instruction and arithmetic expression are
provided together in the user message. For example, the expression
$10-6$ is presented as
\begin{quote}
\texttt{Output ONLY a number. 10 - 6 =}
\end{quote}
before applying the model's instruction chat template. Schematically, the
resulting prompt is
\begin{quote}
\ttfamily\small
<bos><|turn>user\\
Output ONLY a number. 10 - 6 =<turn|>\\
<|turn>model\\
<|channel>thought\\
<channel|>
\end{quote}

For image inputs, the arithmetic expression is rendered inside the image and
the accompanying text contains only the answer-format instruction,
\begin{quote}
\texttt{Output ONLY a number.}
\end{quote}
so the numerical content is available exclusively through the visual input.
After applying the same chat template, the prompt is schematically
\begin{quote}
\ttfamily\small
<bos><|turn>user\\
\texttt{[image]}\\
Output ONLY a number.<turn|>\\
<|turn>model\\
<|channel>thought\\
<channel|>
\end{quote}
where \texttt{[image]} denotes the image tokens inserted by the processor.

\textbf{Datasets.}
Addition uses $a,b\in[0,99]$ with $a+b\leq99$ and subtraction uses
$a,b\in[0,99]$ with $a\geq b$, giving $5050$ expressions each.
Multiplication, used only in \cref{app:multiplication}, uses
$a\in[0,99]$ and $b\in[0,9]$, giving $1000$ expressions. Each record stores
the operands, result, and modular labels $r\bmod 10$ and $r\bmod 100$.

\textbf{Image rendering.}
Image conditions are generated from the same arithmetic records, so operands,
operation, and result are identical across formats. Only the arithmetic
expression is rendered; the answer-format instruction stays in the accompanying
text message. Each expression is drawn as centered black text in 64-pixel DejaVu Sans (Book) font on a white $384\times384$ canvas and stored as PNG. Rendering is
deterministic, with the same font, position, scale, and colours for every
example and no augmentation. Images are passed to the
\texttt{Gemma4-12B-it} \texttt{AutoProcessor} with its default resizing,
normalization, and visual-token settings.

\textbf{Behavioural filtering.}
Unless an experiment studies model errors, analyses use model-correct subsets:
an example is retained when the generated output parses as an integer and
matches the target exactly. Filtering runs separately per format, and matched
cross-modal comparisons use the intersection of examples answered correctly in
every format being compared.

\textbf{Activation positions.}
Activations are read after each transformer block from the
\texttt{resid\_post} hook. For text, positions are defined semantically rather
than by raw tokenizer index. Each semantic element is aligned with its model
tokens, and when an element splits into multiple subtokens we select the final
one, so the recorded activation has processed the complete element.
\Cref{tab:semantic-positions} gives the alignment for the example above.
Textual probe and intervention sweeps use positions $10$--$17$; position $17$
is the final pre-generation \texttt{<channel|>} token and is the common
reference position used in the main analyses.

\begin{table}[t]
\centering
\small
\caption{Example semantic-token alignment for the textual prompt
\texttt{Output ONLY a number. 10 - 6 =}. When a semantic element spans
multiple model tokens, the final subtoken is used.}
\label{tab:semantic-positions}
\begin{tabular}{c l @{\qquad} c l}
\toprule
Position & Semantic token & Position & Semantic token \\
\midrule
0  & \texttt{<bos>}      & 9  & \texttt{-} \\
1  & \texttt{<|turn>}    & 10 & \texttt{6} \\
2  & \texttt{user}       & 11 & \texttt{=} \\
3  & \texttt{Output}     & 12 & \texttt{<turn|>} \\
4  & \texttt{ONLY}       & 13 & \texttt{<|turn>} \\
5  & \texttt{a}          & 14 & \texttt{model} \\
6  & \texttt{number}     & 15 & \texttt{<|channel>} \\
7  & \texttt{.}          & 16 & \texttt{thought} \\
8  & \texttt{10}         & 17 & \texttt{<channel|>} \\
\bottomrule
\end{tabular}
\end{table}

\subsection{Reference layer and position}
\label{app:reference-config}

Text position sweeps cover semantic positions $10$ to $17$. While linear probes reveal strong result decodability at earlier positions, activation patching and \das{} identify position $17$, the final prompt position before generation, as the strongest tested site for causal control. The dedicated \das{} position sweep at layer $43$ confirms the strongest autoregressive intervention accuracy there. This position is also the only one with a common
meaning across modalities: earlier text positions correspond to operands,
operators, and chat-template elements that have no visual counterpart, whereas
the final pre-generation position is the same final channel token in both
formats, reached after the complete input has been processed and with several
blocks remaining before the output readout.

Layer $43$ lies inside the late region where result geometry is stable and
interventions are consistently strong. We use it as a common reference across
conditions rather than claiming every condition peaks exactly there.

\subsection{DAS training and causal dimension}
\label{app:das-training}
\textbf{Training data and intervention pairs.}
Examples are split sample-disjointly into training, validation, and test sets
using fractions $0.70/0.15/0.15$. Intervention pairs are sampled independently
within each split as ordered base--donor pairs. Base and donor examples are
distinct and differ in the target numerical result while satisfying any
experiment-specific matching constraints. Standard runs use at most $4096$
training pairs, $512$ validation pairs, $512$ test pairs, and $128$
autoregressive test pairs. When more valid pairs are available, pairs are
selected by uniform reservoir sampling and shuffled before training.

Given a counterfactual answer \(y^{\mathrm{cf}}\), the DAS basis is optimized using the summed teacher-forced negative log-likelihood over the target tokens, averaged across examples:
\[
\mathcal L_{\mathrm{DAS}}
=
-\frac{1}{B}
\sum_{i=1}^{B}
\sum_{j=1}^{m_i}
\log P\left(
y_{i,j}^{\mathrm{cf}}
\mid x_{\mathrm{base}}^{(i)},y_{i,<j}^{\mathrm{cf}};
h_{\mathrm{base}}^{\prime(i)}
\right).
\]

This loss is used only for training; causal performance is evaluated separately
with interchange intervention accuracy, with $\iiaar$ as the principal metric.

\textbf{Optimization and initialization.}
DAS is optimized with Adam at learning rate $10^{-4}$ and batch size $16$,
with gradient clipping at $1.0$. All runs use PCA-based initialization: PCA is
estimated from up to $3584$ unique training examples and retains the smallest
number of components explaining $90\%$ of the activation variance. The DAS
basis is initialized as a random orthonormal $k$-dimensional subspace inside
this retained PCA span, rather than as the leading $k$ principal components.

Exploratory layer--dimension sweeps use $8$ epochs with early-stopping patience
$2$. The main \(k=22\) experiments instead use three independent seeds, a maximum of $15$ epochs, and early-stopping patience 3.

\begin{figure}[t]
    \centering
    \includegraphics[width=0.72\linewidth]
    {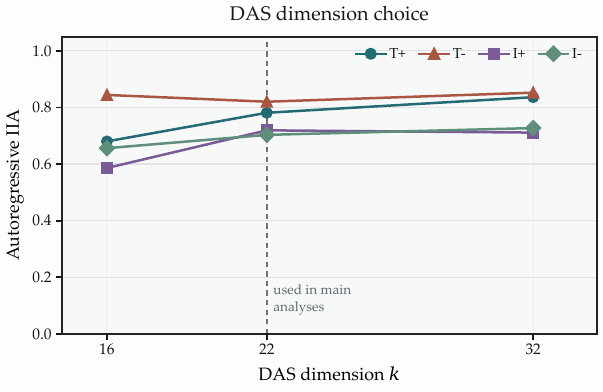}
    \caption{
    \textbf{Epoch-matched DAS dimension selection.}
    Autoregressive IIA for $k\in\{16,22,32\}$ across text and image addition
    and subtraction. Using $k=32$ as the reference, we compare retention
    across the tested dimensions. A dimension of $k=16$ falls below the
    $90\%$ retention threshold in two conditions, whereas $k=22$
    satisfies it across all four. We therefore retain $k=22$ for the
    main analyses, retraining the final subspaces with the longer
    $15$-epoch budget.
    }
    \label{fig:app-das-dimension}
\end{figure}

\textbf{Causal dimension.}
We choose the common DAS dimensionality through an epoch-matched sensitivity
analysis over
\[
k\in\{16,22,32\}
\]
for text and image addition and subtraction. Taking $k=32$ as the reference,
we select the smallest common dimension satisfying
\[
\iiaar(k) \geq 0.90\,\iiaar(32)
\]
in every condition. As shown in \cref{fig:app-das-dimension}, $k=16$ falls below the $90\%$ retention threshold in two conditions, whereas $k=22$ satisfies it in all four conditions.
We therefore use $k=22$ throughout the main analyses.

\clearpage
\section{Numerical geometry in text arithmetic}

This appendix reports the diagnostic characterization of the textual reference
conditions. None of it is required for the cross-condition results, but it
establishes where an organized numerical representation appears and motivates
the choice of comparison object made in \cref{sec:setup}.

\subsection{Emergence and consolidation of a base-10 code}

\begin{figure}[htb]
    \centering
    \includegraphics[width=0.98\textwidth]
        {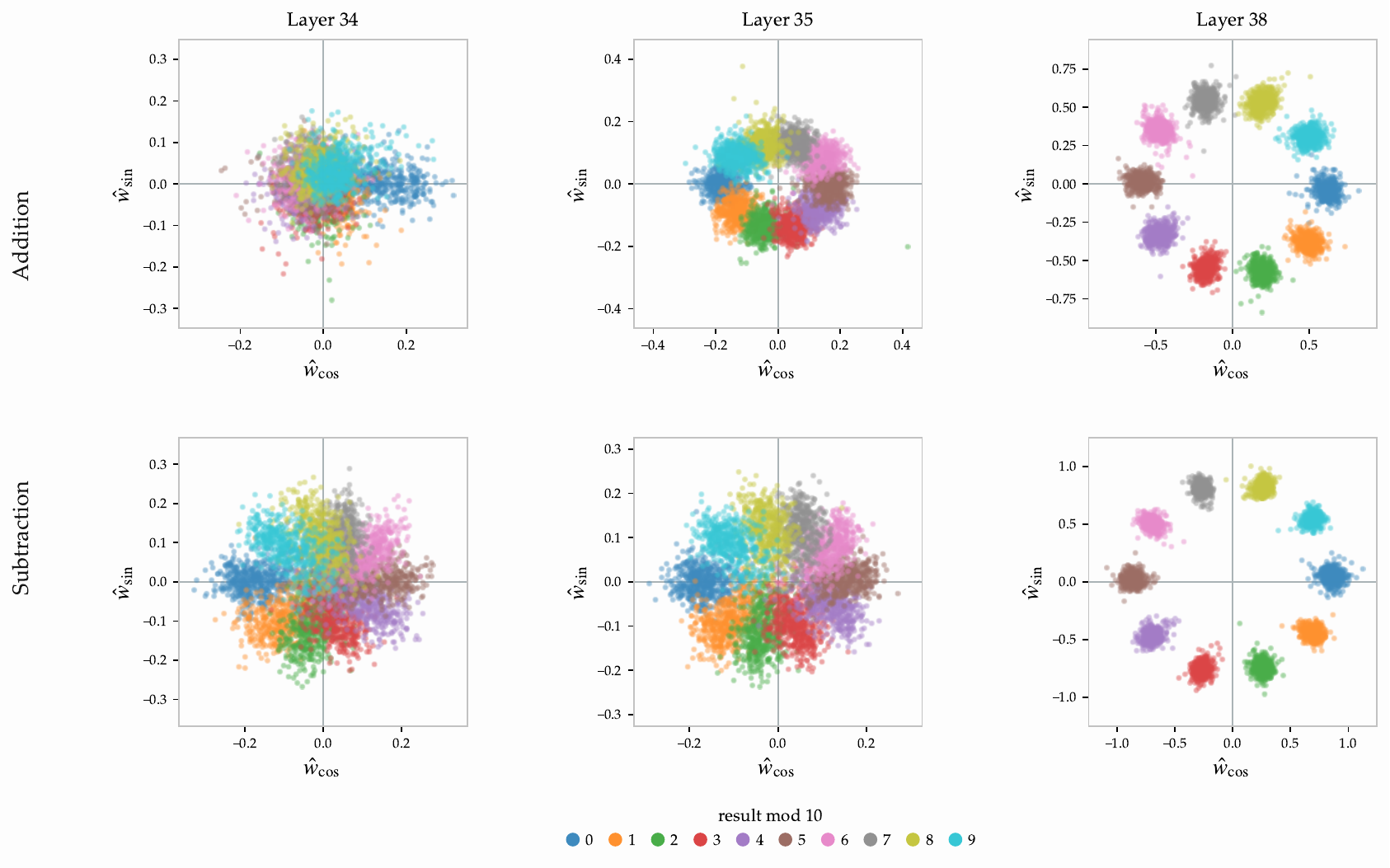}
    \caption{
        \textbf{A base-10 result geometry becomes progressively organized
        across layers.}
        Residual stream activations for addition and subtraction projected onto
        the orthonormalized period-$10$ Fourier plane at layers $34$, $35$ and
        $38$, coloured by result modulo $10$.
    }
    \label{fig:app-fourier-planes}
\end{figure}

For addition and subtraction, a weak result signal exists in earlier layers and
linear probe accuracy rises sharply around layer $35$. Periodic structure
appears at the same depth. \Cref{fig:app-fourier-planes} shows the transition
for the period-$10$ component: little separation by result modulo $10$ at layer
$34$, a circular ordering emerging at layer $35$, and clear structure by layer
$38$.

\begin{figure}[htb]
    \centering
    \includegraphics[width=0.98\textwidth]
        {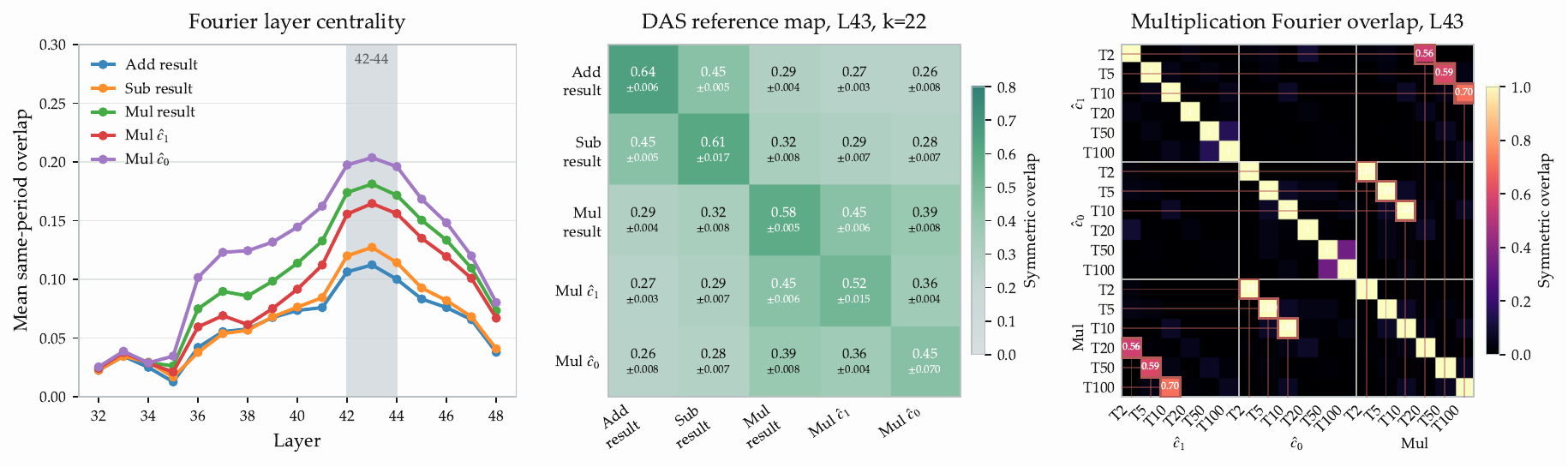}
    \caption{
        \textbf{The numerical representation stabilizes into partially shared,
        operation-dependent late states.}
        \textbf{(Left)} Cross-layer stability of the Fourier representation,
        averaged first over periods $T\in\{5,10,20,50,100\}$ and then over
        comparison layers. Layers $42$ to $44$ achieve the highest mean
        overlap.
        \textbf{(Middle)} Symmetric overlap between independently learned
        $k=22$ \das subspaces. Addition and subtraction share a substantial
        component ($0.45$) while remaining below within-variable consistency
        across seeds. The multiplication variables form a second family.
        \textbf{(Right)} Cross-variable Fourier plane overlap between the
        multiplication intermediates and the final product.
        All panels use activations at semantic position $17$.
    }
    \label{fig:app-overlap}
\end{figure}

First appearance is not the most stable form. Comparing period-matched Fourier
planes across late layers with the symmetric subspace overlap
$\kappa_{\mathrm{sym}}(A,B) = \|Q_A^\top Q_B\|_F^2 / \min(k_A,k_B)$, the planes
at layers $42$ to $44$ have the greatest average overlap with surrounding late
layers (\cref{fig:app-overlap}, left). We therefore distinguish emergence of
the code around layer $35$ from its consolidation into a stable coordinate
system in the low forties, which is the range the main analyses use.

A result probe trained on one operation transfers strongly to the other in both
directions, so the same linear readout stays informative across operations
within the restricted answer space. Their Fourier planes at layers $42$ to $44$
nonetheless align only weakly, an early instance of the pattern that recurs
throughout the paper: compatible information, non-identical coordinates.

\subsection{Multiplication}
\label{app:multiplication}

Multiplication is excluded from the four-condition analyses because its causal
organization is more distributed and changes with depth, not because it lacks
numerical structure. Writing $a=10a_1+a_0$ and $b\in\{0,\ldots,9\}$, we track
the intermediates of \citet{BaiEtAl2025},
$\hat c_0 = a_0 b$, $c_0 = \hat c_0 \bmod 10$, and
$\hat c_1 = a_1 b + \lfloor \hat c_0/10 \rfloor$.

\begin{figure*}[t]
    \centering
    \includegraphics[width=0.48\textwidth]
    {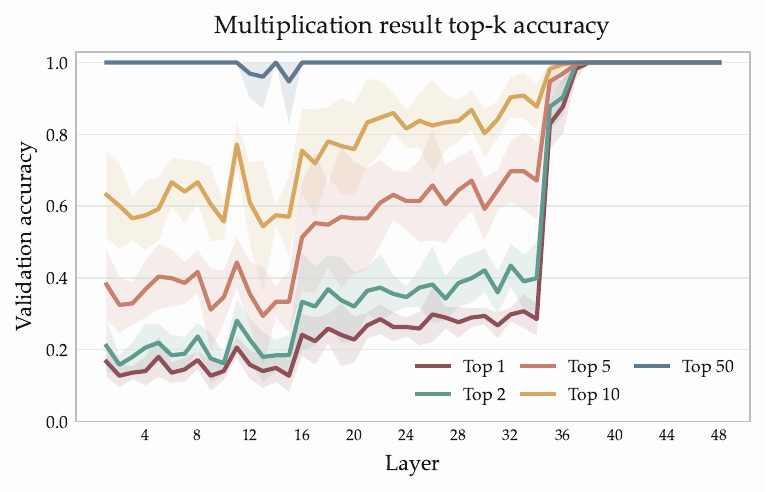}
    \includegraphics[width=0.48\textwidth]
    {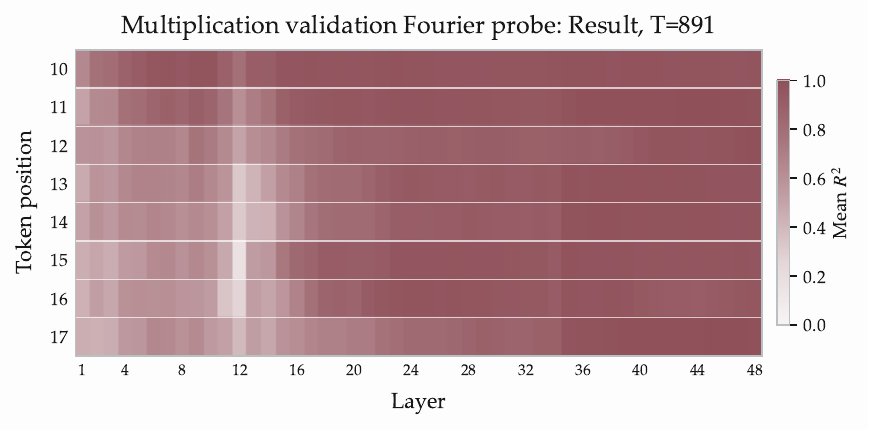}
    \caption{
    \textbf{Multiplication develops from coarse magnitude to exact
    classification.}
    \textbf{Left:} Top-\(k\) accuracy for the result-balanced
    exact-product linear probe. The increase with \(k\) shows that the
    correct product is often restricted to a smaller candidate set
    before becoming the top prediction.
    \textbf{Right:} Held-out mean \(R^2\) for a long-period Fourier
    probe spanning the multiplication result range. The earlier Fourier
    signal is consistent with approximate magnitude information
    preceding exact product classification.
    }
    \label{fig:app-mul-refinement}
\end{figure*}

\textbf{The product is refined from coarse to precise.}
The final product does not become available all at once. On the
result-balanced subset, exact top-1 classification remains weak at depths where
a long-period Fourier probe spanning the product range is already informative.
At the same layers, top-$k$ accuracy rises substantially with $k$: the correct
product is often not yet the most likely class, but has already been restricted
to a smaller candidate set. This is consistent with a coarse numerical estimate
being progressively refined into an exact product
(\cref{fig:app-mul-refinement}).

On a result-balanced subset, exact product decodability rises sharply around
layer $35$ and the late representation develops Fourier structure. The
intermediates relate to the product through decimal place value, and the
Fourier overlap map recovers exactly those relations
(\cref{fig:app-overlap}, right): since $r = c_0 + 10\hat c_1$ we have
$r \bmod T = \hat c_0 \bmod T$ for $T\in\{2,5,10\}$, so $\hat c_0$ shares the
short-period planes of the result, while a period-$T$ component of $\hat c_1$
contributes at period $10T$, so its period-$2$, $5$ and $10$ planes align with
the product's period-$20$, $50$ and $100$ planes.

\begin{figure*}[!t]
    \centering
    \includegraphics[width=0.92\textwidth]
    {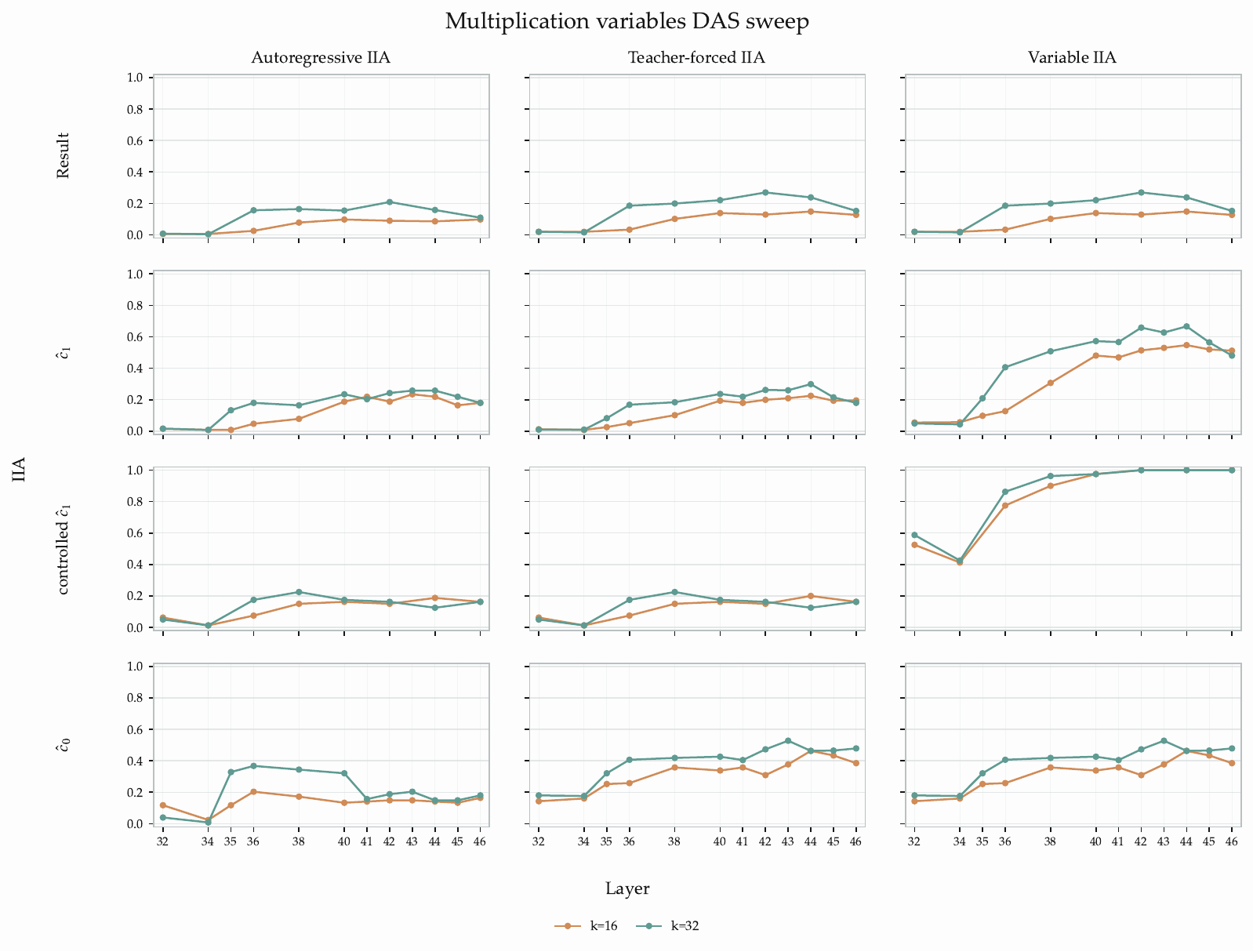}
    \caption{
        \textbf{\das{} layer and dimensionality sweeps for multiplication
        variables.}
        Rows show the final result, \(\hat{c}_1\), controlled
        \(\hat{c}_1\), and \(\hat{c}_0\); columns report autoregressive,
        teacher-forced, and variable-level \iia{} for \(k=16\) and \(k=32\).
        In the controlled \(\hat{c}_1\) condition, base and donor examples
        share \(r \bmod 100\) but differ in \(\hat{c}_1\). The intervention
        achieves near-perfect variable-level control in the late layers,
        but substantially weaker answer-level control.
    }
    \label{fig:app-das-mul}
\end{figure*}

\textbf{The causal organization is more distributed.}
DAS sweeps over the final product and multiplication intermediates confirm that
their causal roles differ across depth (\cref{fig:app-das-mul}). Compared with
addition and subtraction, the final product and its intermediates show weaker
and more distributed autoregressive control. The effect of $\hat c_0$ is more
localized, with a clearer peak around layer $36$, whereas $\hat c_1$ shows a
weaker but more sustained influence across the later layers.

A controlled $\hat c_1$ intervention makes this distinction clearer. We select
pairs whose answers contain at least three digits and satisfy
$r_{\mathrm{base}}\bmod100=r_{\mathrm{donor}}\bmod100$ while differing in
$\hat c_1$, thereby fixing the lower two answer digits while changing the
higher-order multiplication state. Under this pairing, variable-level IIA
becomes nearly perfect in the late-middle layers, while teacher-forced and
autoregressive answer IIA remain substantially lower. Thus, $\hat c_1$ can be
causally manipulated as an internal variable without being sufficient by itself
to reproduce the complete answer.

The causal geometry also changes with depth. At layer $43$, the final-product
DAS space aligns most strongly with $\hat c_1$, while $\hat c_0$ occupies more
distinct coordinates; a $\hat c_0$ space learned at layer $36$ progressively
loses alignment with later multiplication spaces. Multiplication therefore
passes through a connected but evolving family of intermediate numerical
states rather than one stable result space.

\subsection{Fourier geometry is a signature, not a causal basis}
\label{app:fourier}

To relate the visible periodic geometry to the causal subspaces, we measure
directional containment $\kappa(R\leftarrow F) = \|Q_R^\top Q_F\|_F^2 / k_F$ of
a Fourier plane inside the corresponding \das space. Fourier directions are
consistently better represented than random subspaces predict, but containment
is far from complete.

\begin{figure}[htb]
    \centering
    \includegraphics[width=0.98\textwidth]
        {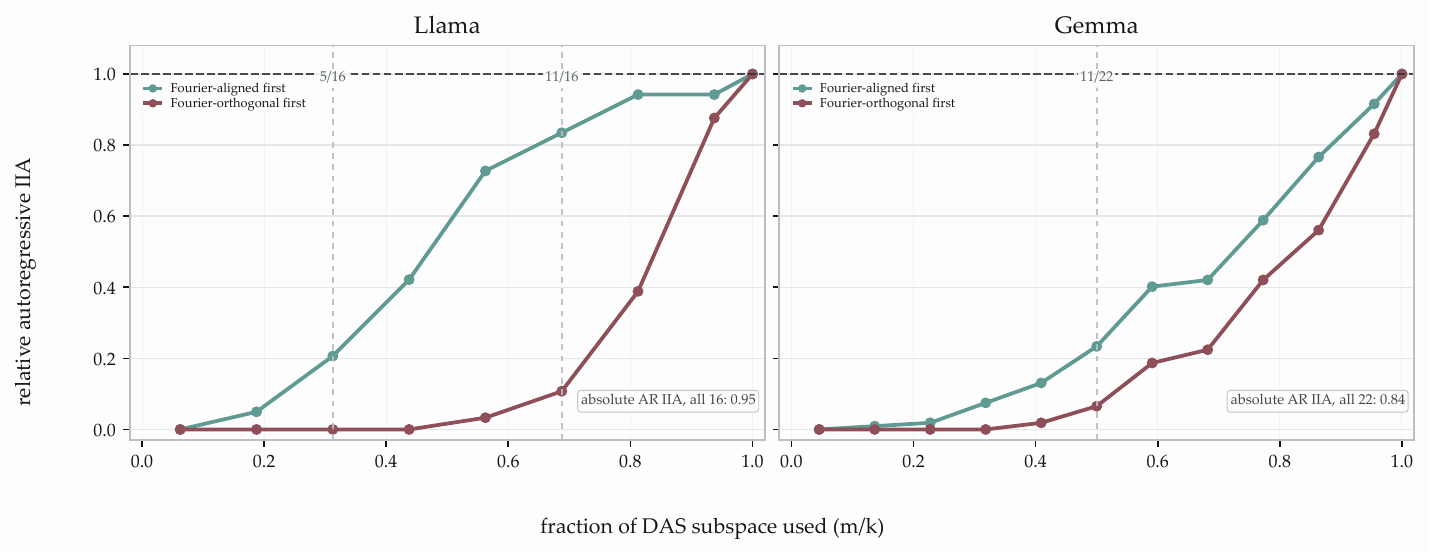}
    \caption{
        \textbf{The causal role of Fourier geometry differs across models.}
        \textbf{(Left)} In \llama, directions most aligned with the Fourier
        subspace recover most of the full \das effect.
        \textbf{(Right)} \gemma is more distributed: Fourier-aligned directions
        recover a minority of the effect and substantial recovery needs
        directions outside the visible Fourier geometry.
        The $x$-axis reports $m/k$, where $m$ is the number of retained \das
        directions ordered by Fourier alignment and $k$ is the full \das
        dimensionality.
    }
    \label{fig:app-fourier-watermark}
\end{figure}

We then order the \das directions by alignment with the Fourier subspace.
Taking $F^\top R = \mU\mSigma\mV^\top$ and $\vq_i = R\vv_i$ gives directions
spanning the same causal space but ordered by principal alignment with Fourier
geometry. Intervening cumulatively from either end
(\cref{fig:app-fourier-watermark}) shows that in \llama the $11$
Fourier-visible directions recover about $84\%$ of the full autoregressive
$\iiaar$ while the remaining five contribute little in isolation, whereas in
\gemma the corresponding $11$ recover about $23\%$, the remaining $11$ about
$7\%$, and the full effect appears only when both are combined. The geometric
alignment differs correspondingly: the mean squared principal cosine over the
Fourier-visible directions is $13.1\%$ in \llama against $2.6\%$ in \gemma.

Consistently, direct Fourier steering in \llama makes the intended value the
top prediction in $52\%$ of interventions, while in \gemma diagonal top-1
accuracy stays near chance ($\approx 5\%$) across the tested result-forming
layers and prompt positions, so the failure is not explained by an
inappropriate intervention site.

\clearpage
\section{Cross-condition transport}

\subsection{Visual arithmetic develops a comparable late causal state}
\label{app:visual}

Image addition and subtraction follow the same representational progression as
their textual counterparts. Results become strongly linearly accessible in the
middle-late layers, periodic structure emerges in the same region, and the
clearest signal sits near the final pre-generation position. Activation
patching and \das identify strong causal influence in the late layers
($42$--$45$) at that position, matching the reference configuration of
\cref{app:reference-config}.

The conditions differ in how controllable they are. Within-condition $\iiaar$
reaches $0.747$ for image addition and $0.693$ for image subtraction,
against $0.831$ and $0.862$ for text addition and subtraction, respectively.
The final pre-generation position remains preferred under the dedicated causal
sweep, although the best layer and the effect of increasing $k$ vary across
conditions. These different condition-specific causal ceilings motivate the
destination-normalized transfer used in the main analyses
(\cref{eq:normalized-transfer}); the corresponding raw $\iiaar$ values and
seed-level diagnostics are reported below.

\subsection{Direct reuse: raw causal effects and controls}
\label{app:direct-transfer}

\begin{figure}[htb]
    \centering
    \includegraphics[width=0.65\textwidth]{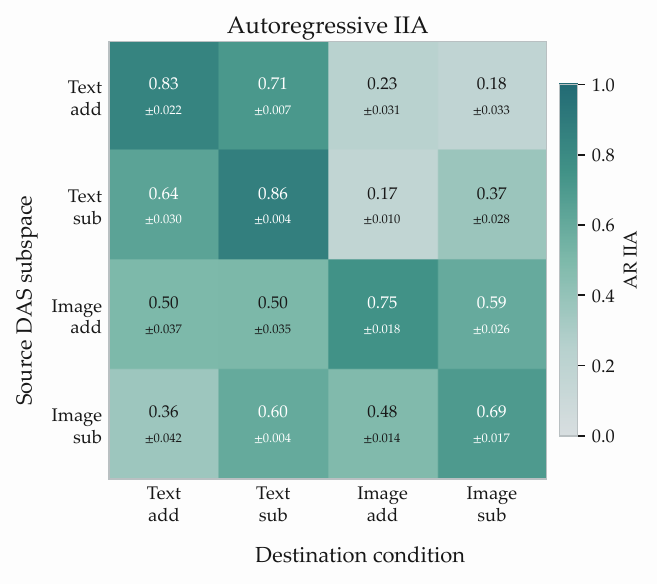}
    \caption{
    \textbf{Arithmetic result subspaces show partial direct causal
    transfer across operations and modalities.}
    Raw $\mathrm{IIA}_{\mathrm{AR}}$ when the subspace learned
    on the source condition is used to intervene on base and
    donor activations drawn from the destination condition.
    Entries report mean $\pm$ standard deviation across the
    $3\times3$ combinations of source and destination DAS seeds.
    }
    \label{fig:app-direct-transfer}
\end{figure}

\Cref{fig:transport}a reports destination-normalized direct reuse, which
accounts for the different causal ceilings of the four destination conditions.
For completeness, \cref{fig:app-direct-transfer} reports the corresponding raw
autoregressive intervention accuracy before normalization.

The control subspace in \cref{eq:normalized-transfer} is a random
$k$-dimensional subspace sampled from the retained PCA span of the destination
activations and evaluated on the same base--donor pairs, which keeps the
baseline equal in dimension and restricted to directions the destination
actually occupies.

The random subspace supplies the normalization baseline for all four
destinations. Each same-operation cross-modal direction is evaluated over all
$3\times3$ independently trained source--destination DAS seed combinations
(\cref{tab:app-direct-seeds}).

\begin{table}[t]
\centering
\scriptsize
\caption{\textbf{Per-seed destination-normalized direct transfer.}
Rows denote source DAS seeds and columns destination DAS seeds. Each
$3\times3$ block reports one same-operation cross-modal transfer direction.}
\label{tab:app-direct-seeds}
\resizebox{\linewidth}{!}{%
\begin{tabular}{c|ccc|ccc|ccc|ccc}
\toprule
& \multicolumn{3}{c|}{$\Tadd\!\to\!\Iadd$}
& \multicolumn{3}{c|}{$\Tsub\!\to\!\Isub$}
& \multicolumn{3}{c|}{$\Iadd\!\to\!\Tadd$}
& \multicolumn{3}{c}{$\Isub\!\to\!\Tsub$} \\
Source seed
& 0 & 1 & 2
& 0 & 1 & 2
& 0 & 1 & 2
& 0 & 1 & 2 \\
\midrule
0 & .306 & .323 & .312 & .535 & .517 & .505 & .582 & .610 & .615 & .691 & .691 & .685 \\
1 & .245 & .258 & .250 & .512 & .494 & .484 & .527 & .552 & .558 & .700 & .700 & .694 \\
2 & .337 & .355 & .344 & .605 & .584 & .571 & .627 & .657 & .663 & .691 & .691 & .685 \\
\bottomrule
\end{tabular}}
\end{table}

\begin{figure*}[!t]
    \centering
    \includegraphics[width=0.65\textwidth]
    {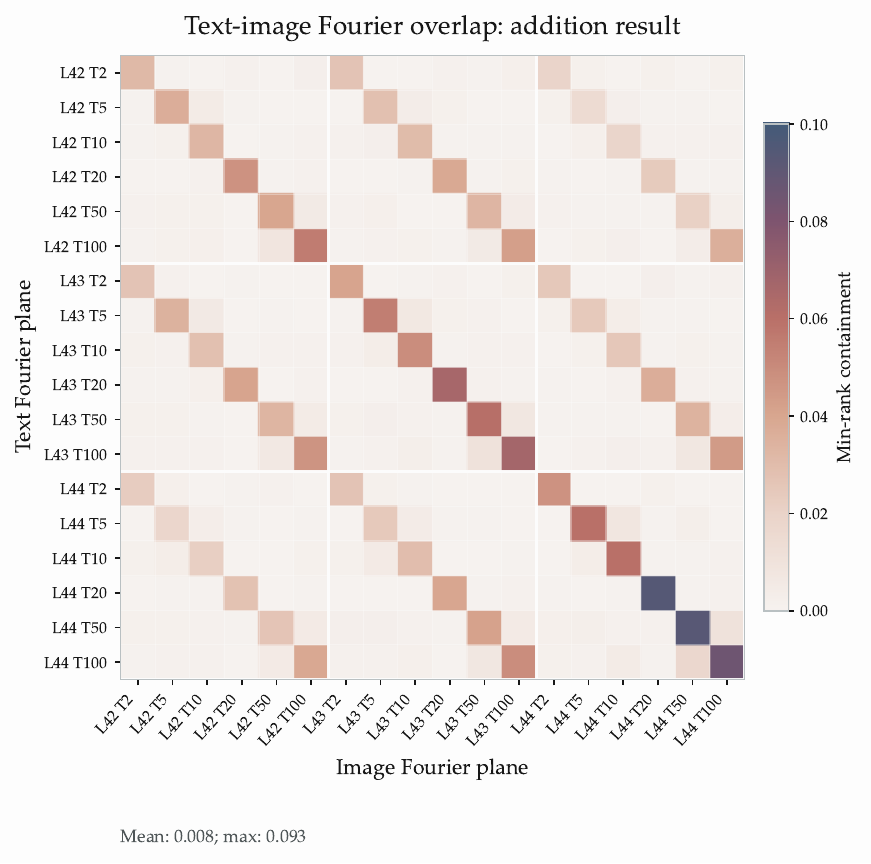}
    \caption{
    \textbf{Cross-modal alignment of result Fourier planes.}
    Min-rank containment between textual and visual addition-result Fourier
    planes across layers and periods. Cross-modal alignment remains weak,
    including between matched periods, despite the periodic numerical
    structure visible separately within each modality.
    }
    \label{fig:fourier-overlap-image-text}
\end{figure*}

Fourier planes also align only weakly across modalities. For image addition,
mean text--image min-rank containment over layers $42$--$44$ and the tested
periods is $0.0076$, with a maximum of $0.093$ for the matched layer-$44$,
period-$20$ planes (\cref{fig:fourier-overlap-image-text}). Thus, although both
modalities independently develop periodic numerical structure, those visible
Fourier planes do not form one common cross-modal basis.

\subsection{Choice of alignment target}
\label{app:alignment-targets}

\Cref{tab:app-alignment-variants} compares three quantities one could align:
complete \das coordinates, result-conditioned centroids, and the base-to-donor
displacement that the interchange intervention actually modifies. A rotation
fitted on displacements already improves mean normalized recovery to $0.718$,
compared with $0.532$ under direct reuse. Adding a global scale further raises
recovery to $0.914$.

\begin{table}[htb]
    \centering
    \caption{
        \textbf{Comparison of cross-modal alignment targets.}
        Destination-normalized transfer, averaged over the $3\times3$ source
        and destination \das seed combinations.
    }
    \label{tab:app-alignment-variants}
    \begin{tabular}{lcccc}
        \toprule
        Method & $\Tadd\!\to\!\Iadd$ & $\Tsub\!\to\!\Isub$
               & $\Iadd\!\to\!\Tadd$ & $\Isub\!\to\!\Tsub$ \\
        \midrule
        Direct transfer                   & 0.303 & 0.534 & 0.599 & 0.692 \\
        Absolute-state $\mQ$              & 0.546 & 0.772 & 0.708 & 0.804 \\
        Centroid-displacement $\mQ$       & 0.531 & 0.795 & 0.705 & 0.816 \\
        Displacement $\mQ$                & 0.531 & 0.811 & 0.704 & 0.828 \\
        Scaled displacement $\alpha\mQ$   & 0.791 & 0.983 & 0.950 & 0.932 \\
        \bottomrule
    \end{tabular}
\end{table}

\subsection{Transformation ablations}
\label{app:transport-ablations}

\Cref{tab:app-transport-ablation} isolates the contributions of orientation and
magnitude to same-operation cross-modal transport. The fitted orthogonal map
$Q$ alone improves mean normalized recovery from $0.532$ to $0.718$, while
adding a global scale raises it further to $0.914$. Applying the fitted scale
without the learned rotation produces essentially no causal transfer; random
orthogonal maps also fail, with or without scaling. The learned scaled
orthogonal map $\alpha Q$ recovers $0.791$, $0.983$, $0.950$ and $0.932$ of
the destination-specific effect across the four directions. Allowing a fully
unrestricted linear map reaches $0.963$ mean recovery, showing that most
of the usable linear correspondence is captured by one learned orientation and
one global scale.

\begin{table}[htb]
    \centering
    \caption{
        \textbf{Full scaled-displacement ablation.}
        Destination-normalized transfer, averaged over the $3\times3$ seed
        combinations.
    }
    \label{tab:app-transport-ablation}
    \begin{tabular}{lcccc}
        \toprule
        Method & $\Tadd\!\to\!\Iadd$ & $\Tsub\!\to\!\Isub$
               & $\Iadd\!\to\!\Tadd$ & $\Isub\!\to\!\Tsub$ \\
        \midrule
        Direct                                  & 0.303 & 0.534 & 0.599 & 0.692 \\
        Displacement $\mQ$                      & 0.531 & 0.811 & 0.704 & 0.828 \\
        Scale only $\alpha\mI$                  & 0.004 & 0.004 & 0.003 & 0.012 \\
        Random rotation $\mQ_{\mathrm{rand}}$   & 0.001 & 0.004 & 0.000 & 0.001 \\
        Scaled random $\alpha\mQ_{\mathrm{rand}}$ & 0.001 & 0.011 & 0.002 & 0.004 \\
        Full $\alpha\mQ$                        & 0.791 & 0.983 & 0.950 & 0.932 \\
        Unrestricted linear $\mA$               & 0.889 & 0.993 & 1.001 & 0.970 \\
        \bottomrule
    \end{tabular}
\end{table}

\subsection{Rank-restricted transport}
\label{app:rank-restricted}

How many directions does the correspondence actually require? If only a handful
carried it, cross-modal alignment would reduce to a small set of dominant axes.
Let $X$ and $Y$ contain matched source and destination training
displacements. We decompose their cross-covariance as
\[
X^\top Y = V\Sigma U^\top
\]
and order the matched directions by decreasing singular value. We use the
cross-covariance rather than the fitted orthogonal map $Q$, whose singular
values are all one and therefore provide no ordering. Retaining the leading
$m$ matched pairs gives
\[
\Delta z_B^{(m)}
=
\alpha U_mV_m^\top\Delta z_A.
\]

\begin{figure*}[!t]
    \centering
    \includegraphics[width=0.8\textwidth]
    {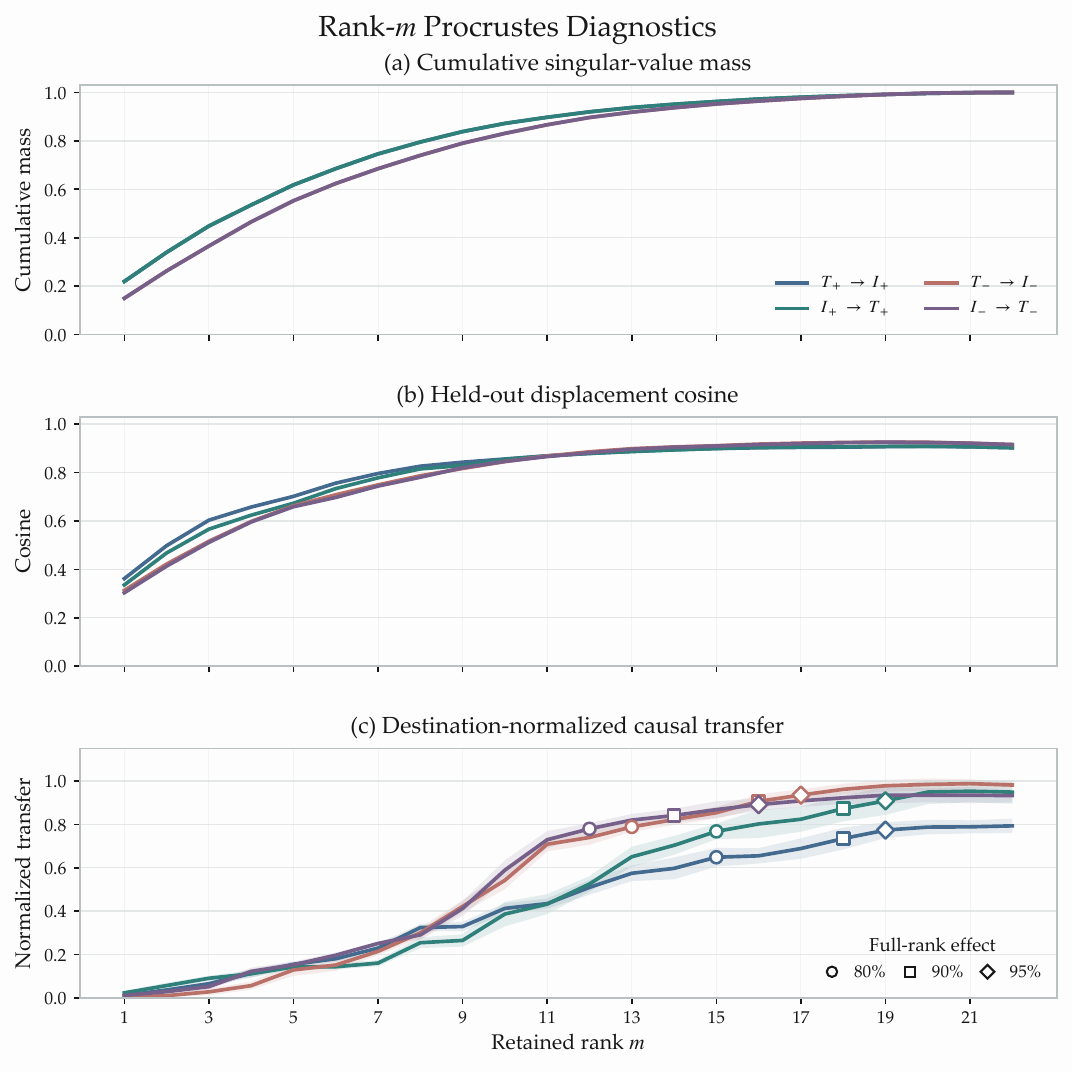}
    \caption{
        \textbf{Full diagnostics for cumulative rank-\(m\) transport.}
        From top to bottom: cumulative singular-value mass, cosine
        alignment on test expressions, and destination-normalized
        causal transfer. Geometric quantities
        saturate earlier than causal performance, showing that the
        leading matched modes are not sufficient to reproduce the full
        intervention effect.
    }
    \label{fig:app-rank-sweep-full}
\end{figure*}

The full sweep separates three quantities: cumulative singular-value mass,
held-out displacement cosine, and destination-normalized causal transfer
(\cref{fig:app-rank-sweep-full}). Geometry saturates substantially earlier than
causal control. The number of modes required to reach $95\%$ of the full-map causal
recovery is $19$, $17$, $19$ and $16$ for $\Tadd\to\Iadd$, $\Tsub\to\Isub$,
$\Iadd\to\Tadd$ and $\Isub\to\Tsub$, respectively.

The directions that best summarize the cross-condition correspondence geometrically
are therefore not sufficient to reproduce it causally. This cumulative ordering
shows that the leading modes alone are insufficient; it should not be
interpreted as measuring the isolated causal contribution of each individual
mode.

\subsection{Cross-operation and composed transports}
\label{app:cross-operation}

When the arithmetic operation changes, expressions no longer correspond
one-to-one. An addition and a subtraction expression may represent the same
numerical change while sharing neither operands nor computation. We therefore
match the condition-specific spaces by the ordered numerical transition implemented by the
intervention. For condition $A$, we define
\[
\Delta z^{A}_{a\rightarrow b}
=
\mathbb{E}\!\left[
\Delta z^{A}
\mid
y_{\mathrm{base}}=a,\,
y_{\mathrm{donor}}=b
\right],
\]
so $a\rightarrow b$ always denotes changing the represented result from $a$ to
$b$, regardless of which expressions realize those values.

Scaled orthogonal maps are fitted between these transition-conditioned means,
with one mean vector contributing per ordered transition. Applying the same
estimator to the easier same-operation cross-modal relations, where exact
expression matching is also available, provides a control for estimator choice.
Averaging over expressions may remove expression-specific information while
preserving the numerical transition geometry.

The fitted transport is also effective when only the operation changes:
mean normalized recovery reaches $0.828$, compared with $0.773$ for direct
reuse (\cref{fig:transport}a). These means come from different pair-sampling
protocols and are not a paired comparison. Thus, operation changes already
admit strong causal reuse without alignment, while fitted maps also preserve
their numerical transition geometries. Relations changing both factors reach
$0.807$ mean normalized recovery (\cref{tab:app-cross-operation-direct}).

\begin{table}[t]
\centering
\small
\caption{
\textbf{Direct transports that change both modality and operation.}
Each map is fitted directly between the indicated source and destination
condition using matched ordered numerical transitions. Entries report
mean $\pm$ sample standard deviation of destination-normalized autoregressive
transfer across the $3\times3$ source--destination DAS seed combinations.
}
\label{tab:app-cross-operation-direct}
\begin{tabular}{lc}
\toprule
Transport & Normalized transfer \\
\midrule
$\Iadd \rightarrow \Tsub$ & $0.847 \pm 0.057$ \\
$\Isub \rightarrow \Tadd$ & $0.842 \pm 0.082$ \\
$\Tadd \rightarrow \Isub$ & $0.839 \pm 0.047$ \\
$\Tsub \rightarrow \Iadd$ & $0.699 \pm 0.069$ \\
\midrule
Mean & $0.807$ \\
\bottomrule
\end{tabular}
\end{table}

We next ask whether a relation that changes both factors can be reconstructed
without fitting that relation directly. For each such source--destination pair,
we compare the direct map with two paths: one changes modality first and
operation second, and the other reverses the order. For example,
$\Tadd\!\rightarrow\!\Isub$ can be reached through
\[
\Tadd \rightarrow \Iadd \rightarrow \Isub
\]
or through
\[
\Tadd \rightarrow \Tsub \rightarrow \Isub.
\]
The two composed paths preserve almost all of the direct causal effect
(\cref{tab:app-composed-paths}). Averaged over the four relations, direct
transport reaches $0.807$ destination-normalized transfer, compared with
$0.799$ for modality-first and $0.799$ for operation-first composition.

\begin{table}[t]
\centering
\small
\caption{
\textbf{Causal performance of direct and composed transport paths.}
Entries report mean $\pm$ sample standard deviation of
destination-normalized autoregressive transfer across the $3\times3$
source--destination DAS seed combinations. The composed paths use only the
two simpler pairwise maps along the corresponding
modality-first or operation-first route.
}
\label{tab:app-composed-paths}
\begin{tabular}{lccc}
\toprule
Transport & Direct & Modality first & Operation first \\
\midrule
$\Iadd \rightarrow \Tsub$
& $0.847 \pm 0.057$
& $0.845 \pm 0.043$
& $0.846 \pm 0.049$ \\

$\Isub \rightarrow \Tadd$
& $0.842 \pm 0.082$
& $0.841 \pm 0.071$
& $0.835 \pm 0.074$ \\

$\Tadd \rightarrow \Isub$
& $0.839 \pm 0.047$
& $0.817 \pm 0.037$
& $0.833 \pm 0.043$ \\

$\Tsub \rightarrow \Iadd$
& $0.699 \pm 0.069$
& $0.695 \pm 0.074$
& $0.682 \pm 0.083$ \\
\midrule
Mean
& $0.807$
& $0.799$
& $0.799$ \\
\bottomrule
\end{tabular}
\end{table}

\begin{figure*}[!t]
    \centering
    \includegraphics[width=0.94\textwidth]
    {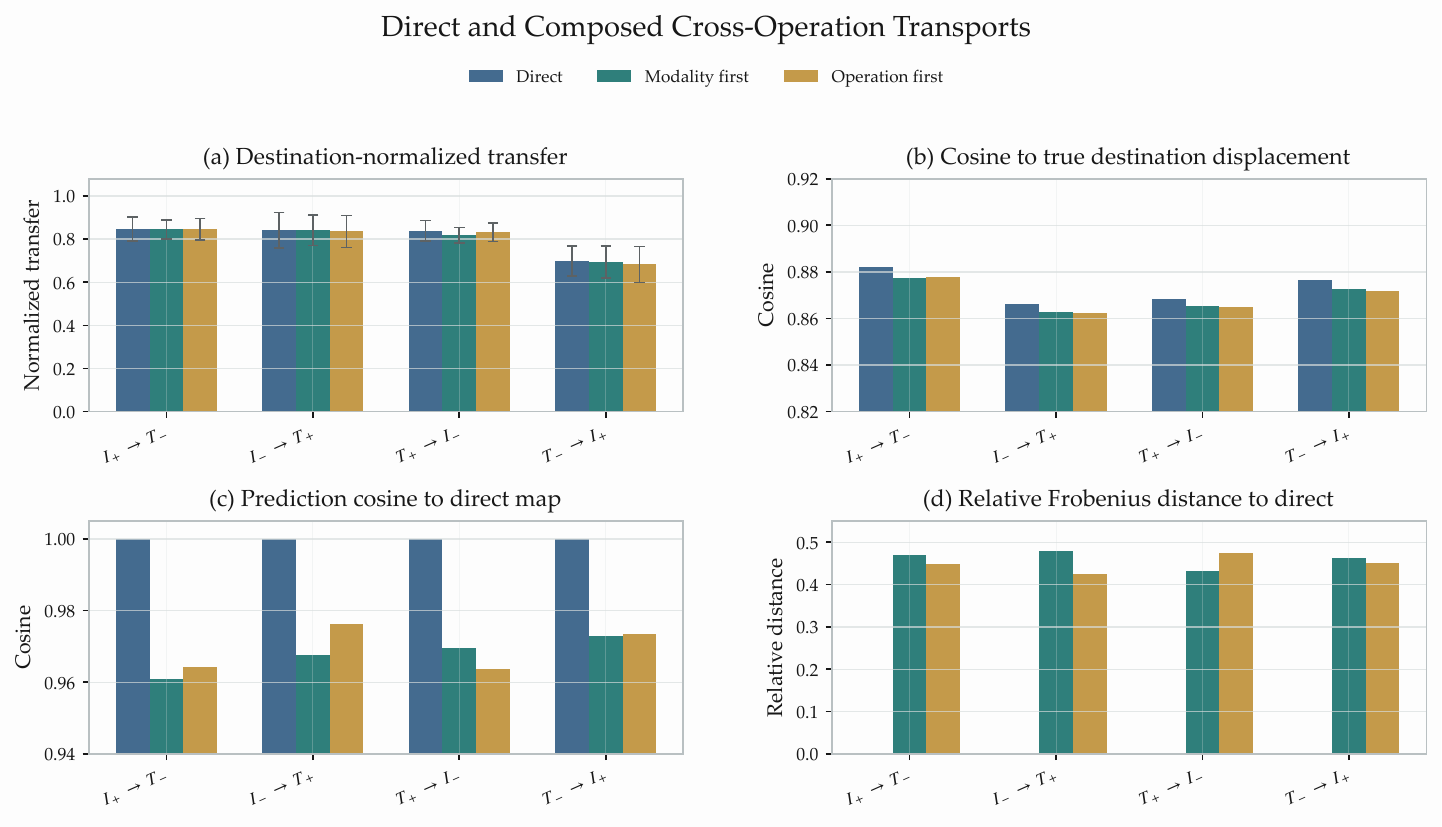}
    \caption{
        \textbf{Complete composed-path comparison.}
        For each cross-operation direction, the figure compares the
        directly fitted transport with the modality-first and
        operation-first compositions. Panels report destination-normalized
        causal transfer, cosine to the true destination displacement,
        cosine between the composed and direct-map predictions, and
        normalized Frobenius distance to the direct operator.
    }
    \label{fig:app-factorized-paths-full}
\end{figure*}

The geometric comparison shows where this agreement holds
(\cref{fig:app-factorized-paths-full}). On held-out numerical transitions, the
displacements predicted by the modality-first and operation-first paths have
mean cosine $0.968$ and $0.969$, respectively, with those predicted by the
directly fitted maps. Agreement on the observed numerical transitions does not,
by itself, imply equality between the full $22$-dimensional transformations.

\clearpage
\section{Readout decomposition and the common frame}

\subsection{Digit-readout overlap and ablation controls}
\label{app:digit-readout}

The centered digit-discriminative span $U_{\mathrm{D}}$ defined in
\cref{sec:readout} has rank $9$. For an orthonormal DAS basis
$R\in\mathbb{R}^{d_{\mathrm{model}}\times22}$, we quantify its coupling to this
readout as
\[
E_{\mathrm{D}}
=
\frac{
\left\|U_{\mathrm{D}}^\top R\right\|_F^2
}{
\|R\|_F^2
}.
\]
Learned DAS spaces are substantially more aligned with the digit readout than
random subspaces of the residual stream: for text addition at layer $43$, the
observed containment is $41.65$ times the analytic random residual-subspace
expectation. Untrained PCA and random-in-PCA controls show $22.78$ and $23.90$
times this expectation, respectively, indicating additional readout enrichment
in the trained causal DAS spaces.

\begin{figure}[htb]
    \centering
    \includegraphics[width=0.95\textwidth]{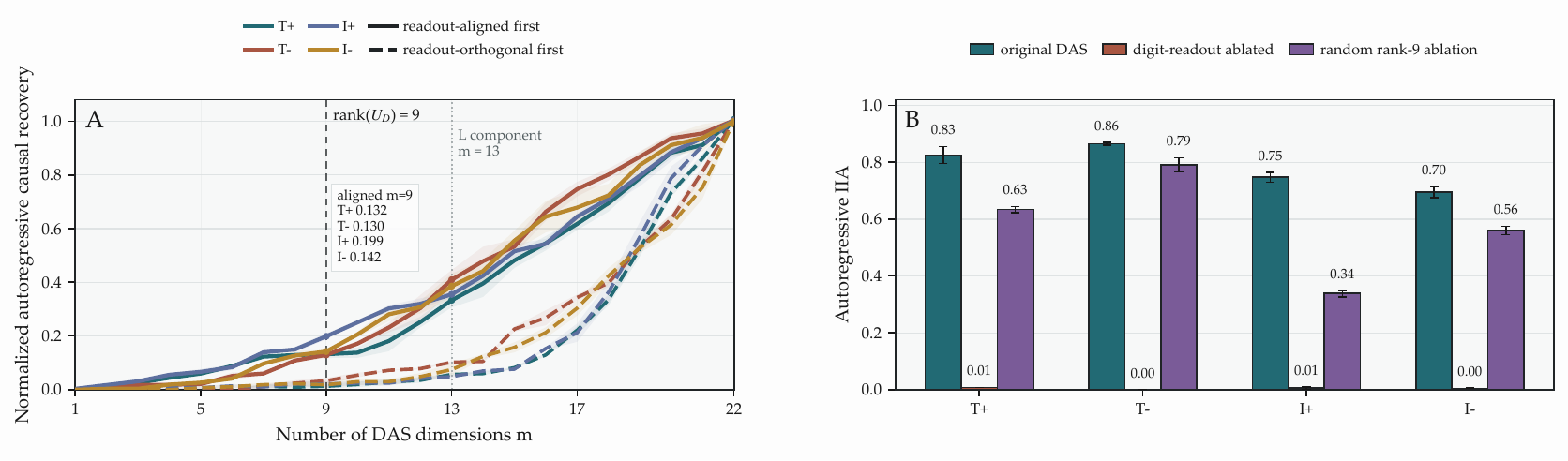}
    \caption{
        \textbf{Final-position causal control depends specifically on
        digit-readout coupling, but is carried by neither component alone.}
        \textbf{(a)} Cumulative autoregressive recovery with the $22$ \das
        directions ordered from most to least digit-readout aligned, and in
        reverse. The dashed line marks the rank-$9$ boundary between $\Dsp$ and
        $\Lsp$ defined in \cref{sec:readout}. Neither component is sufficient by
        itself.
        \textbf{(b)} Removing $\Udig$ from the complete DAS state almost eliminates
        autoregressive control in all four conditions, whereas ablating a
        dimension-matched random span preserves a substantial fraction of the
        original effect.
    }
    \label{fig:app-readout}
\end{figure}

Across the four arithmetic conditions, approximately $9.8\%$--$11.0\%$ of the
squared DAS basis norm lies along the digit-readout span. Removing this component
nevertheless almost eliminates autoregressive control. Mean self-$\iiaar$ falls
from $0.826$, $0.865$, $0.747$ and $0.695$ for $\Tadd$, $\Tsub$, $\Iadd$ and
$\Isub$ to $0.008$, $0.000$, $0.005$ and $0.003$, respectively. The effect is
specific to the readout geometry rather than to removing nine dimensions: a
dimension-matched random rank-$9$ ablation leaves $0.634$, $0.791$, $0.339$ and
$0.560$ (\cref{fig:app-readout}b).

These values are evaluated on the $128$-pair held-out readout-ablation set;
the resulting full-DAS baselines differ slightly from those in the main
evaluation. The learned spaces, layer, and intervention position are unchanged.

As an independent functional check that this coupling reflects a reusable
digit-output component rather than arithmetic-specific structure, we apply the
text-addition DAS intervention to a non-arithmetic one-digit repeat task, such
as
\begin{quote}
\texttt{Repeat this number: 7 =}
\end{quote}
which requires the same digit output vocabulary but no arithmetic computation.
The arithmetic-trained space reaches $0.489$ teacher-forced IIA,
whereas random and untrained PCA-based controls reach zero. Under the stricter
eight-token complete-answer autoregressive criterion, transfer is only $1.5\%$. Thus, part of
the final-position causal state learned from arithmetic can be reused directly
to control the immediate digit output even when no arithmetic computation is
required.

\subsection{Readout-free geometry}
\label{app:readout-free}

\textbf{Exact decomposition.}
For each learned DAS basis $R$, we compute
\[
U_D^\top R = A\Sigma V^\top
\]
and rotate the basis within its own span as
\[
\widetilde R
=
RV
=
[\tilde r_1,\ldots,\tilde r_{22}].
\]
This changes only the coordinates inside the learned space, since
\[
\widetilde R\widetilde R^\top = RR^\top.
\]
The overlap matrix has numerical rank $9$ for every condition and DAS seed, so
we define
\[
D=[\tilde r_1,\ldots,\tilde r_9],
\qquad
L=[\tilde r_{10},\ldots,\tilde r_{22}].
\]
Across conditions and seeds,
\[
\|U_D^\top L\|_F = O(10^{-7}),
\qquad
\|\Pi_D+\Pi_L-RR^\top\|_F = O(10^{-6}),
\]
so $L$ lies in the exact readout null space within the learned DAS
representation and the decomposition discards no part of the original causal
subspace.

\textbf{Value-held-out evaluation.} All readout-free geometry experiments use a value-disjoint split. The $100$
possible result values are divided into $80$ alignment-training values and
$20$ held-out values. Writing
\[
\mu_i(y)=\mathbb{E}[z_i\mid \mathrm{result}=y],
\qquad
\Delta_i(a\rightarrow b)=\mu_i(b)-\mu_i(a),
\]
a transition contributes to the fitted scaled orthogonal map only when both
$a$ and $b$ belong to the $80$ training values. Evaluation uses exclusively
transitions whose endpoints belong to the $20$ held-out values. No numerical
value appearing during evaluation therefore contributes to the fitted
cross-condition transformation; this holdout is during alignment fitting,
not during the preceding DAS training.

Because displacement alignment does not determine the origin of each coordinate
system, one condition-specific offset is estimated from the training-value
centroids and held fixed during retrieval of the unseen values. Top-1 retrieval
is among the $20$ held-out candidates, so chance is $5\%$.

\Cref{tab:app-readout-free} compares readout-orthogonal representations and
controls. Projecting the digit span out of the complete DAS representation
leaves mean held-out transition cosine $0.908$ and $98.1\%$ top-1 retrieval.

The exact $13$-dimensional $\Lsp$ component retains most of the relational
structure while giving up some ability to identify individual values:
transition cosine remains $0.807$, with $75.5\%$ top-1 retrieval. A
dimension-matched random readout-orthogonal subspace reaches $0.558$
transition cosine and $38.3\%$ top-1, while shuffling destination value identities
before fitting reduces transition cosine to $-0.003$ and retrieval to
$5.2\%$ (approximately chance). Thus, $\Lsp$
concentrates substantially more cross-condition numerical structure than a
generic readout-orthogonal slice, and the alignment depends on matching the
same numerical identities across conditions.

\textbf{No-fit RSA.} As a fitting-free check, Representational Similarity Analysis compares the
pairwise distance matrices of the held-out value centroids without estimating
any cross-condition transformation:
\[
d_i(a,b)=\|\mu_i(a)-\mu_i(b)\|_2.
\]
The mean cross-condition Spearman correlation is $0.808$ after projecting
the digit span out of the complete DAS representation and $0.765$ in the
exact $\Lsp$ spaces, compared with $0.545$ for dimension-matched random
readout-orthogonal subspaces. For $\Lsp$, a $1000$-permutation numerical-label test reaches
\[
p=\frac{1}{1001},
\]
the minimum attainable under this procedure. The readout-free correspondence
is therefore visible even before any cross-condition map is fitted.

\begin{table}[t]
\centering
\small
\caption{
\textbf{Cross-condition numerical geometry after separating the current digit
readout.}
Transition cosine and retrieval are evaluated only on numerical values excluded
from the alignment fit. Top-1 retrieval is among $20$ held-out values, so chance
is $5\%$. A dash indicates that RSA was not evaluated.
}
\label{tab:app-readout-free}
\begin{tabular}{lccc}
\toprule
Representation & Transition cosine & Top-1 & RSA $\rho$ \\
\midrule
Projected DAS (22d)
    & $0.908$ & $98.1\%$ & $0.808$ \\
Exact $\Lsp$ (13d)
    & $0.807$ & $75.5\%$ & $0.765$ \\
Random readout-orthogonal (13d)
    & $0.558$ & $38.3\%$ & $0.545$ \\
Shuffled value correspondence
    & $-0.003$ & $5.2\%$ & -- \\
\bottomrule
\end{tabular}
\end{table}

\FloatBarrier
\subsection{Synchronization protocol}
\label{app:sync-protocol}

All pairwise $\Lsp$-space maps are fitted from training-value transitions only.
Full-graph evaluation reconstructs every relation from the four synchronized
frames. Leave-one-relation-out (LORO) evaluation removes both directions of one
relation before synchronization and reconstructs it from the remainder.
Because synchronization is fitted only to displacements, it does not determine
the origin of the condition coordinate systems. Condition-specific offsets are
therefore estimated separately from training-value centroids and used only for
placing numerical values; they do not contribute to the synchronized rotations
or scales.

Synchronization is performed independently for each matched DAS seed, so that
each graph contains one coherent set of four condition-specific spaces. The
direct-pairwise reference reported here is therefore restricted to matched
source--destination seeds; this accounts for the small difference from the
$0.807$ mean in \cref{sec:readout}, which averages all $3\times3$ seed
combinations.

\begin{table}[htb]
    \centering
    \caption{
        \textbf{Global consistency of the exact $\Lsp$ spaces.}
        All metrics are evaluated on numerical values excluded from the
        pairwise alignment fits.
    }
    \label{tab:app-L-sync}
    \begin{tabular}{lccc}
        \toprule
        Map & Transition cosine & Top-1 & Top-5 \\
        \midrule
        Direct pairwise         & 0.801 & 73.4\% & 98.4\% \\
        Full synchronization    & 0.791 & 74.8\% & 98.9\% \\
        LORO synchronization    & 0.762 & 71.1\% & 98.3\% \\
        Modality-only frames    & 0.249 & 18.1\% & 55.3\% \\
        \bottomrule
    \end{tabular}
\end{table}

A stricter two-frame control ties addition and subtraction within each modality,
$\mU_{\Tadd}=\mU_{\Tsub}$ and $\mU_{\Iadd}=\mU_{\Isub}$, with scales tied
likewise. It drops to transition cosine $0.249$ and $18.1\%$ top-1. The four-frame result
therefore cannot be reduced to one text--image transformation. LORO also
preserves substantial unseen-value retrieval ($71.1\%$ top-1).

\subsection{Held-out-condition additive closure}
\label{app:heldout-condition}

The leave-one-condition-out experiment tests whether the modality and operation
effects observed in three corners of the $2\times2$ design can predict the
numerical geometry of the fourth. This differs from synchronization:
synchronization asks whether observed pairwise transformations are jointly
consistent with one set of condition frames, whereas here one complete condition
is absent from the prediction.

Let $i^\star$ denote the held-out condition, $r$ its diagonally opposite
observed condition, and $m$ and $o$ the two remaining conditions, differing from
$r$ in modality and operation respectively. We use $r$ as the reference
coordinate system. For any observed condition $i$, let
\[
\mathcal A_{i\rightarrow r}(z)
=
\alpha_{r\leftarrow i}Q_{r\leftarrow i}z
+
b_{r\leftarrow i}
\]
denote its affine alignment into the reference coordinates, where
$Q_{r\leftarrow i}$ and $\alpha_{r\leftarrow i}$ are fitted from
training-value displacements and $b_{r\leftarrow i}$ is the offset estimated
from the training-value centroids.

For each numerical value $v$, the missing representation is predicted as
\[
\widehat z_{i^\star}(v)
=
z_r(v)
+
\Bigl(
\mathcal A_{m\rightarrow r}(z_m(v))-z_r(v)
\Bigr)
+
\Bigl(
\mathcal A_{o\rightarrow r}(z_o(v))-z_r(v)
\Bigr).
\]
The two differences are respectively the modality and operation changes
observed relative to the reference condition.

For example, when $\Isub$ is held out, $\Tadd$ is the reference and
\[
\widehat z_{\Isub}(v)
=
z_{\Tadd}(v)
+
\Bigl(
\mathcal A_{\Iadd\rightarrow\Tadd}(z_{\Iadd}(v))
-z_{\Tadd}(v)
\Bigr)
+
\Bigl(
\mathcal A_{\Tsub\rightarrow\Tadd}(z_{\Tsub}(v))
-z_{\Tadd}(v)
\Bigr).
\]
The true $\Isub$ representations do not contribute to this prediction. They
are aligned into the same reference coordinates only afterwards, for
evaluation using an anchor map estimated for that purpose. Evaluation values are disjoint from those used to estimate the
pairwise alignments.

We repeat the construction with each of $\Tadd$, $\Tsub$, $\Iadd$ and $\Isub$
held out in turn, across three matched DAS seeds and three value splits, giving
$36$ leave-one-condition-out evaluations. Averaged over the four held-out
conditions, the predicted geometry reaches held-out transition cosine $0.629$,
with $58.9\%$ top-1 retrieval among the $20$ unseen values.

As a correspondence control, we repeat the construction after permuting the
numerical identities of the same-modality neighbouring condition before forming
the prediction, using four independent shuffles per configuration. Transition
cosine falls to $-0.109$ and top-1 retrieval to $3.8\%$, approximately chance.

The experiment should therefore be interpreted as an approximate additive
closure of modality and operation effects in aligned $\Lsp$ coordinates, not
as exact separability of those effects. No autoregressive intervention is
performed on the predicted held-out $\Lsp$ geometry.

\subsection{A local successor relation transfers across conditions}
\label{app:successor-equivariance}

\begin{figure}[t]
    \centering
    \includegraphics[width=\columnwidth]
    {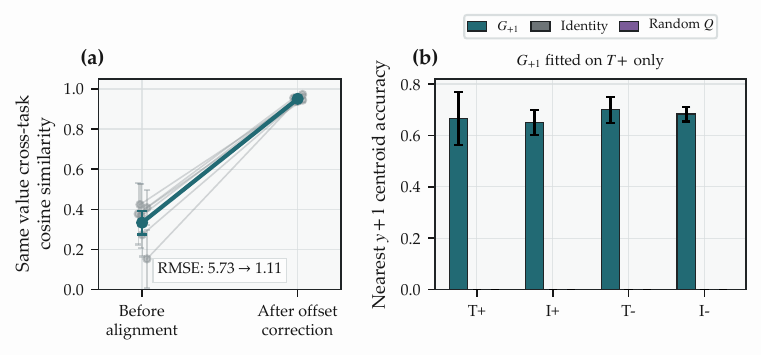}
    \caption{
    \textbf{A local successor relation transfers across the shared
    numerical frame.}
    After registering the four full $22$-dimensional DAS spaces into common
    coordinates, an orthogonal successor operator \(G_{+1}\) is fitted using text
    addition only and frozen. On held-out numerical values, the same
    operator moves representations toward the corresponding \(y+1\)
    centroid in text/image addition and subtraction, while identity and
    random-orthogonal controls fail. Transfer is strongest for one local
    step and degrades under repeated application.
    }
    \label{fig:app-successor}
\end{figure}

The common frame predicts more than correspondence between the same
numerical values. If it captures a shared numerical geometry, a relation
learned between nearby values in one condition should remain meaningful
in the others. We test this with the simplest such relation: the
successor transformation \(y\mapsto y+1\).

We first register the complete $22$-dimensional DAS spaces with synchronized
orientations and scales and offsets estimated from training-value centroids,
analogously to \cref{eq:frame-absolute} but using full DAS coordinates rather
than the exact $13$-dimensional $\Lsp$. Write $u_i^{(22)}(y)$ for a
registered numerical centroid in condition $i$. Using text addition only, we
fit an orthogonal operator
\[
G_{+1}^\star
=
\argmin_{G^\top G=I}
\sum_{y\in\mathcal{Y}_{\mathrm{train}}}
\left\|
u_{\Tadd}^{(22)}(y+1)-G u_{\Tadd}^{(22)}(y)
\right\|_2^2.
\]
The numerical values used for evaluation are excluded entirely from the
edges used to fit or validate the operator. We then freeze
\(G_{+1}^\star\) and evaluate it without refitting on all four
conditions.

The same local relation transfers across modality and operation
(\Cref{fig:app-successor}). The transformed representation
\(G_{+1}u_i^{(22)}(y)\) is nearest to the correct \(y+1\) centroid in
\(66.7\%\) of text-addition cases, \(65.0\%\) of image-addition cases,
\(70.0\%\) of text-subtraction cases, and \(68.3\%\) of
image-subtraction cases. Identity and random-orthogonal controls remain
near zero. As a behavioural check, the same transformation raises the
teacher-forced top-\(5\) presence of the entire successor target from \(0.758\)
under the identity control to \(0.887\) on average. Complete-answer
autoregressive accuracy under the operator is only \(9.5\%\), however.

A transformation learned from one corner of the \(2\times2\) design
therefore preserves the same local numerical relation in the other
three. This gives a direct, local view of the common geometry established
by the pairwise and synchronized analyses above: what transfers is not
only numerical identity, but also how nearby numerical identities are
related.

The effect is approximate rather than an arithmetic mechanism.
Repeated application of \(G_{+1}\) becomes progressively less accurate
for larger shifts, and the intervention does not reliably make \(y+1\)
the greedy autoregressive answer. We therefore interpret \(G_{+1}\) as
a local regularity of the registered causal numerical geometry, not as evidence
that the model performs arithmetic by repeatedly applying a linear
successor operator.

\subsection{Full-DAS synchronization as a causal control}
\label{app:full-das-sync}

As a behavioural counterpart to the readout-orthogonal synchronization, we
repeat the same procedure in the original $22$-dimensional DAS spaces, before
separating their digit-readout coupling. Unlike $\Lsp$, these complete spaces
support strong autoregressive interventions and therefore allow the synchronized
relations to be evaluated causally.

Across the twelve directed relations, independently fitted pairwise transports
reach mean $\iiaar=0.6385$, while transports reconstructed from one synchronized
orientation and scale per condition reach $0.6359$. On the numerical transitions
occupied by the data, their predicted displacements have mean cosine $0.989$.

Because synchronization is fitted to displacements, it does not determine the
absolute origin of each condition coordinate system. Estimating one
condition-specific offset from training-value centroids raises mean same-value
cross-condition cosine on held-out values from $0.334$ to $0.951$, while
reducing RMSE from $5.73$ to $1.11$. Individual post-registration cosines
can be as low as $0.922$.

This experiment tests whether the fitted full-space transports share a global
organization; it does not imply that fitting improves every individual relation,
since some individual relations may differ (\cref{sec:transport}). We
treat it as supporting evidence rather than the primary shared-frame result
because the complete DAS spaces also contain the digit-readout coupling isolated
in \cref{sec:readout}.

\clearpage
\section{Autoregressive intervention details}

\subsection{Matched and mismatched $\Dsp/\Lsp$ interventions}
\label{app:latent-swap}

The causal interaction experiment of \cref{sec:autoregressive} uses individual
model activations rather than result-conditioned centroids. For a base state
$h_b$ and a donor state $h_y$ representing counterfactual value $y$, we write
the two components of the standard DAS edit as
\[
\Delta h_{\Dsp}^{b\rightarrow y}
=
\Pi_{\Dsp}(h_y-h_b),
\qquad
\Delta h_{\Lsp}^{b\rightarrow y}
=
\Pi_{\Lsp}(h_y-h_b).
\]
Since
\[
\Pi_{\Dsp}+\Pi_{\Lsp}=\mR_i\mR_i^\top,
\]
taking both components from the same donor reconstructs exactly the original
DAS interchange intervention,
\[
h_{\mathrm{match}}
=
h_b
+
\Delta h_{\Dsp}^{b\rightarrow y}
+
\Delta h_{\Lsp}^{b\rightarrow y}.
\]

To separate their roles, we choose a second counterfactual value
$\tilde y\neq y$ and make the two components disagree:
\[
h_{\mathrm{mis}}
=
h_b
+
\Delta h_{\Dsp}^{b\rightarrow y}
+
\Delta h_{\Lsp}^{b\rightarrow \tilde y}.
\]
The matched and mismatched states therefore contain the same readout-coupled
contribution, while only the numerical identity supplied through $\Lsp$ changes.
We additionally evaluate $\Dsp$-only, $\Lsp$-only, and random-latent controls.

\begin{figure}[htb]
    \centering
    \includegraphics[width=0.98\textwidth]{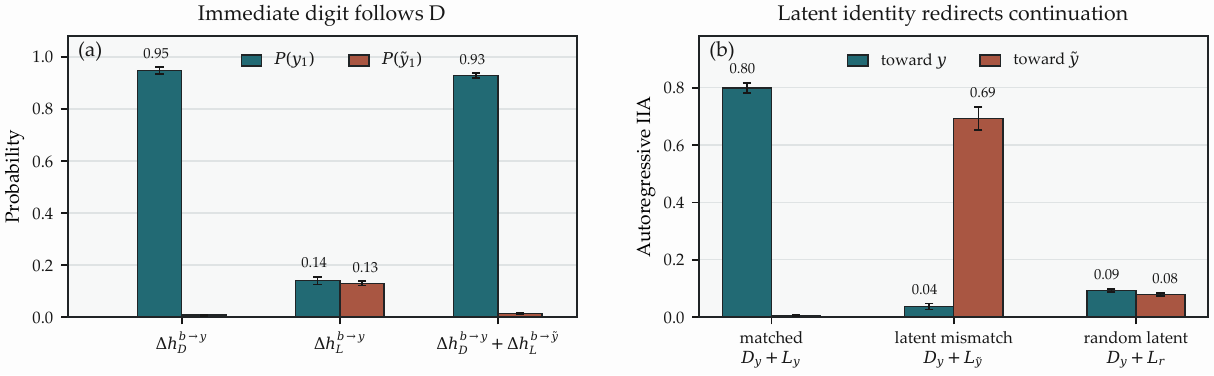}
    \caption{
    \textbf{The readout-coupled and readout-orthogonal components have distinct
    temporal causal roles.}
    \textbf{(a)} \emph{The immediate digit follows $\Dsp$.}
    When the values supplying $\Dsp$ and $\Lsp$ require different first digits,
    the current prediction follows the value inserted through $\Dsp$; an
    $\Lsp$-only intervention does not comparably redirect that digit.
    \textbf{(b)} \emph{Numerical identity in $\Lsp$ redirects the continuation.}
    For values $y$ and $\tilde y$ sharing their first digit but differing in their
    second, the matched intervention recovers $y$, whereas replacing only $\Delta h_{\Lsp}^{b\rightarrow y}$ with
    $\Delta h_{\Lsp}^{b\rightarrow\tilde y}$  redirects autoregressive behaviour toward
    $\tilde y$. A random latent displacement does not reproduce this
    value-specific effect.
    }
    \label{fig:app-latent-swap}
\end{figure}

\Cref{fig:app-latent-swap} separates two effects. First, when $y$ and
$\tilde y$ require different first digits, the immediate prediction follows
the value supplying $\Dsp$: a $\Dsp$-only intervention strongly redirects the
current digit, whereas $\Lsp$ alone does not, and a mismatch remains biased
towards the digit specified by $\Dsp$.

Second, to isolate the later role of $\Lsp$, we restrict to two-digit donor
values
\[
y = 10y_1+y_2,
\qquad
\tilde y = 10y_1+\tilde y_2,
\qquad
y_2\neq\tilde y_2.
\]
The two states now agree on the digit required immediately and disagree only on
the numerical continuation. Across the four conditions and three DAS seeds,
the matched intervention generates $y$ with mean $\iiaar=0.799$. Replacing
only $\Lsp_y$ by $\Lsp_{\tilde y}$ reduces $\iiaar$ toward $y$ to $0.037$
and redirects generation toward $\tilde y$ with $\iiaar=0.692$. A random
latent displacement reaches only $0.089$ toward the selected alternative
value.

Crucially, this redirection leaves the shared current digit essentially
unchanged: its mean probability is $0.963$ in the matched state and $0.965$ in
the mismatch. Numerical identity in $\Lsp$ therefore has a value-specific
causal effect on the continuation despite having no direct coupling to the
current digit readout.

The same-first-digit analysis contains $1009$ matched/mismatched triplets across the
four conditions and three DAS seeds. The random-latent control uses five
displacements per triplet from unrelated numerical activations, excluding the
selected base and counterfactual identities, so it preserves the magnitude and
statistics of a natural latent edit without carrying the selected alternative
value. Metrics are first aggregated within each condition--seed configuration
and then averaged across the twelve configurations.

\textbf{Pairs with different first digits.}
The analysis above uses pairs that share the tens digit. \Cref{tab:app-both-digits}
reports pairs where $y$ and $\tilde y$ differ in both digits ($1041$ triplets per
condition, \eg $y=52$ and $\tilde y=67$). In the mismatched state, $\Dsp$ points
to the tens digit of $y$ and $\Lsp$ to the units digit of $\tilde y$, so the
expected output is the hybrid $57$. The swap produces this hybrid in $51.6\%$ of
cases and reproduces $y$ in only $4.7\%$. For reference, $\Dsp$ alone yields the
hybrid in $7.5\%$ of cases, and the matched edit reproduces $y$ in $74.5\%$, the
ceiling of the $22$-dimensional intervention. The swap thus reaches about
two thirds of this ceiling. The units digit follows $\Lsp$ in $57.7\%$ of swap
trials (matched: $77.9\%$; $\Dsp$ alone: $8.7\%$; chance: $10\%$), while the
tens digit follows $\Dsp$ in $83.9\%$ (matched: $88.3\%$). The gap to the ceiling
is therefore concentrated in the units digit. The random-latent control also
yields a hybrid, with the units digit of the unrelated value, in $49.1\%$ of
cases. The units digit therefore follows whichever value $\Lsp$ carries. The
random control supplies a wrong value and tests whether the selected value
matters. It does not leave the units digit unedited.

\begin{table}[htb]
\centering
\caption{\textbf{Mismatched edits on pairs with different first digits.}
Percentage of generations equal to $y$, to the hybrid (tens digit of $y$ and units
digit of the value supplying $\Lsp$, which is $\tilde y$ except for Random $\Lsp$),
and to the base answer. For Matched, $\Dsp$ only and $\Lsp$ only, the hybrid column
counts how often the output coincides with the hybrid defined from the trial's
$y$ and $\tilde y$.}
\label{tab:app-both-digits}
\begin{tabular}{lccc}
\toprule
Condition & $y$ & Hybrid & Base \\
\midrule
Matched          & 74.5 & 1.7  & 7.8  \\
Swap $\Lsp$      & 4.7  & 51.6 & 11.8 \\
$\Dsp$ only      & 10.4 & 7.5  & 17.4 \\
$\Lsp$ only      & 4.4  & 1.2  & 75.9 \\
Random $\Lsp$    & 8.7  & 49.1 & 10.9 \\
\bottomrule
\end{tabular}
\end{table}

\subsection{Cache propagation protocol}
\label{app:cache-propagation}

The temporal experiment uses the same-first-digit matched and mismatched
construction of \cref{sec:autoregressive,app:latent-swap}, with separately
sampled triplets. The only intervention is applied at layer $43$ of the final pre-generation position $t_0$. After
inserting the matched or mismatched state, the forward pass continues normally
through the remaining layers and the resulting key--value cache is retained.

The shared first digit
\[
y_1=\tilde y_1
\]
is then supplied as the next token in both conditions, and position $t_1$ is
processed using the retained cache with no additional hidden-state
intervention. Because this token is the natural first digit of both candidate
values, matched and mismatched computations receive exactly the same valid
autoregressive prefix. Any difference at $t_1$ must therefore arise from the
earlier intervention propagating through the model's normal cached
computation.

The intervention is introduced after layer $43$ at $t_0$, so cache entries
computed through that layer cannot yet contain intervention-dependent
information. The first downstream entries that can carry the edit appear after
this boundary, matching the sharp rise in conversion cosine from layer $44$ to
layer $45$ reported in \cref{sec:autoregressive}. The analysis uses $1536$
separately sampled triplets across the four conditions and three DAS seeds.

For the layerwise conversion analysis, we project both the mismatch-induced and
natural continuation changes into digit-readout coordinates,
\[
\Delta \vr_\ell^{\mathrm{mis}}
= \Udig^\top(\vh_{\ell,t_1}^{\mathrm{mis}}-\vh_{\ell,t_1}^{\mathrm{match}}),
\qquad
\Delta \vr_\ell^{\mathrm{nat}}
= \Udig^\top[\vh_{\ell,t_1}^{\mathrm{nat}}(\tilde y)-\vh_{\ell,t_1}^{\mathrm{nat}}(y)],
\]
and define the conversion cosine
\[
\gamma_\ell
= \cos(\Delta \vr_\ell^{\mathrm{mis}},\Delta \vr_\ell^{\mathrm{nat}}).
\]
It therefore measures whether the effect carried forward from the earlier
latent edit points in the same digit-facing direction as the model's natural
change from $y$ to $\tilde y$ at $t_1$.

As a value-specificity control, we repeat the propagation using an $\Lsp$
displacement from an unrelated numerical value that shares the same required
first digit. This preserves the valid common prefix while removing the selected
alternative numerical identity. The probability of its alternative units digit
rises only to $0.020$, compared with $0.737$ for the value-specific mismatch.
No intervention is performed at $t_1$ in any condition.

\subsection{Repeat-number comparison}
\label{app:repeat-number}

The repeat-number experiment asks whether the numerical identity in $\Lsp$ is
specific to an arithmetic result or reflects a more general numerical state
associated with a number the model is preparing to produce. We therefore use a
task with no arithmetic computation, for example
\begin{quote}
\texttt{Output ONLY a number. Repeat this number: 43 =}
\end{quote}
where the numerical value is supplied directly.

No repeat-specific DAS space is trained and no intervention is performed.
Instead, the layer-$43$ final-position activation from each repeat prompt is
projected directly into each arithmetic $\Lsp_i$ basis and compared with
arithmetic states representing the same numerical values. The broader analysis
covers values $0,\ldots,99$, with two-digit values used for the principal
geometric comparison.

\begin{figure}[htb]
    \centering
    \includegraphics[width=0.98\textwidth]
        {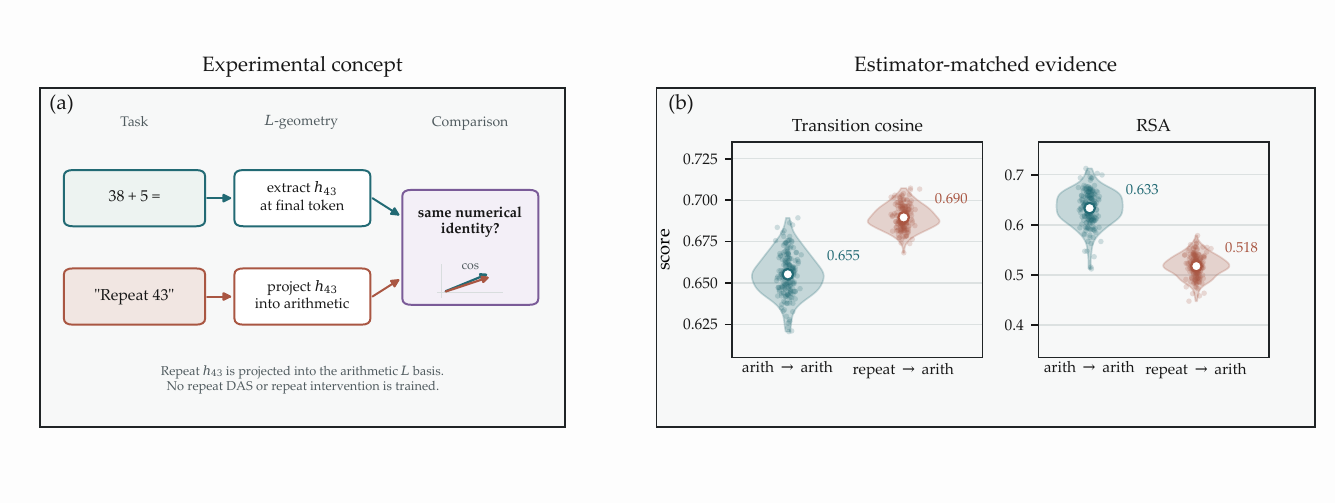}
    \caption{
        \textbf{The readout-orthogonal numerical geometry extends beyond
        arithmetic.}
        \textbf{(a)} The repeat task supplies the numerical value directly.
        No repeat-specific DAS space is trained and no intervention is
        performed; the layer-$43$ final-position activation is projected into
        the arithmetic $\Lsp$ basis and compared with arithmetic states
        representing the same values.
        \textbf{(b)} Estimator-matched comparison over $200$ samples of one
        arithmetic expression per numerical value. Violin distributions show
        held-out transition cosine and no-fit RSA for arithmetic-to-arithmetic
        and repeat-to-arithmetic comparisons, with markers indicating the mean.
    }
    \label{fig:app-repeat}
\end{figure}

A direct comparison has an estimator mismatch: an arithmetic value centroid
averages several expressions producing the same result, whereas the repeat task
contains one prompt per value. We therefore repeat the analysis after sampling
exactly one correctly answered arithmetic expression for each
\[
y\in\{10,\ldots,89\},
\]
and repeat this sampling procedure $200$ times
(\cref{fig:app-repeat}). This removes the advantage arithmetic receives
from averaging multiple examples into each centroid.

Under this estimator-matched comparison, arithmetic-to-arithmetic alignment
reaches mean held-out transition cosine $0.655$, compared with $0.690$ for
repeat-to-arithmetic alignment. Top-1 retrieval is $57.9\%$ against $65.1\%$,
respectively. The fitting-free relational comparison instead favors arithmetic,
with RSA $0.633$ for arithmetic-to-arithmetic and $0.518$ for
repeat-to-arithmetic. The repeat advantage in fitted transition geometry and
the arithmetic advantage in no-fit RSA persist across all $200$ samples.

The repeat task produces the correct answer for all values $0,\ldots,99$
when greedy generation is evaluated at the target answer length, including both
digits of two-digit numbers. Thus, the geometric differences cannot be attributed
to incorrect numerical repetition under this criterion.

The repeat states therefore retain substantial numerical organization
found in arithmetic $\Lsp$: fitted alignment and retrieval can even outperform
arithmetic-to-arithmetic comparisons, although the no-fit geometries differ.
We consequently interpret $\Lsp$ conservatively as a \emph{latent
numerical result state}: a structured representation of numerical identity
reached by several arithmetic routes and, to a substantial degree, by a
non-arithmetic task that requires producing the same number.

\clearpage
\section{SECOND-MODEL REPLICATION IN MINISTRAL}
\label{app:ministral}

The main experiments use a single multimodal model so that text and image
representations can be compared in one residual stream. We therefore ask
whether the central chain of results survives in a second multimodal
architecture. We use \texttt{mistralai/Ministral-3-14B-Instruct-2512-BF16},
which has 40 decoder blocks and residual width $d_{\mathrm{model}}=5120$, and
repeat the same $2\times 2$ comparison between text/image input and
addition/subtraction, denoted $T_+$, $T_-$, $I_+$ and $I_-$. Thinking is
disabled and the output contains digits only. The ten decimal digits are
single output tokens, and two-digit answers are emitted as two successive
digit tokens.

The goal is not to reproduce every diagnostic from the primary model. Instead,
we test the sequence of claims that is central to the paper: whether a late
causal numerical state can be found; whether its coupling to the current digit
readout can be separated exactly from a readout-orthogonal remainder; whether
that remainder preserves numerical geometry across modality and operation;
whether the pairwise relations admit one synchronized frame; and whether the
readout-free numerical identity is used causally during later autoregressive
continuation.

\subsection{Localization and a causal numerical subspace}
\label{app:ministral-causal}

\paragraph{Prompts and chat template.}
We use the model's Hugging Face chat template with
\texttt{use\_chat\_template=true} and thinking disabled. For text arithmetic,
the user message contains the answer-format instruction and the expression
together; for example,
\begin{quote}
\texttt{Output ONLY a number. 10 - 6 =}
\end{quote}
before applying \texttt{add\_generation\_prompt=true}. For image arithmetic,
the arithmetic expression is supplied only through the image and the text
message contains
\begin{quote}
\texttt{Output ONLY a number.}
\end{quote}
with image content placed before text content. Under the current Ministral
Hugging Face template, the corresponding user segments are schematically
\texttt{[INST]...[/INST]} for text and
\texttt{[INST][IMG]...[/INST]} for images. In
both modalities, the final \texttt{[/INST]} token is therefore the common
pre-generation position.

\paragraph{Datasets and image rendering.}
We use the same arithmetic generator and rendering procedure as Appendix~A:
only the arithmetic expression is rendered, as centered black text on a white
$384\times384$ RGB canvas with font size 64. The answer-format instruction is
kept in the accompanying text message. The final datasets are nevertheless
model-specific rather than row-for-row copies of the Gemma datasets, because
examples are filtered separately by Ministral correctness. The saved DAS metadata contains 5,028 retained addition examples in each modality, 5,027 text-subtraction examples, and 5,010 image-subtraction examples.

\paragraph{DAS optimization.}
The final runs use Adam with learning rate $10^{-4}$, batch size 1 and global
gradient clipping at norm 1.0. As in the primary-model pipeline, the basis is
initialized as a random orthonormal $k$-dimensional subspace inside a
train-set PCA span. For Ministral, PCA is estimated from up to 1024 training
activations and retains the smallest span explaining 90\% of activation
variance. Activations are taken from the \texttt{resid\_post} hook. Final runs
use 4096/512/512 training/validation/test intervention pairs, 128
autoregressive evaluation pairs, a 15-epoch budget, early-stopping patience 3,
and three independent DAS seeds.

\paragraph{Layer and position.}
The common intervention site is the final pre-generation \texttt{[/INST]}
token, stored as position $-1$. Activation patching localizes the strongest
common late effect to layers 35--40. At the final prompt position, normalized
teacher-forced patch recovery for $T_+$ is $0.543$ at layer 37 and $0.554$ at
layer 39; for $I_+$ it is $0.484$ and $0.544$, respectively. A direct
$k=32$ DAS comparison shows no autoregressive disadvantage at layer 37:
IIA$_{\mathrm{AR}}=0.188$ at both layers 37 and 39 for $T_+$, and $0.250$ at
both layers for $I_+$. We therefore use layer 37, which retains the late causal
effect while leaving three transformer blocks downstream of the intervention.

\paragraph{Dimension.}
We use $k=32$ as main causal dimension. In full-budget runs at layer
37, increasing to $k=64$ changes IIA$_{\mathrm{AR}}$ from $0.312$ to $0.344$
for $T_+$ and from $0.312$ to $0.336$ for $I_+$, so $k=32$ retains $90.9\%$
and $93.0\%$ of the corresponding $k=64$ effects while halving the dimension.
A lightweight sweep does not show a useful additional gain from $k=64$ to
$k=128$. We therefore fix $k=32$ for all four conditions; this also leaves a
23-dimensional readout-free component after the rank-9 decomposition below.

\begin{table}[t]
\centering
\small
\begin{tabular}{lcccc}
\toprule
Condition & TF-IIA, seeds 0/1/2 & TF mean $\pm$ std &
AR-IIA, seeds 0/1/2 & AR mean $\pm$ std \\
\midrule
$T_+$ & .326/.326/.324 & $.325\pm.001$ &
.312/.297/.289 & $.299\pm.012$ \\
$T_-$ & .500/.504/.486 & $.497\pm.009$ &
.375/.414/.414 & $.401\pm.023$ \\
$I_+$ & .318/.299/.318 & $.312\pm.011$ &
.312/.297/.305 & $.305\pm.008$ \\
$I_-$ & .473/.459/.469 & $.467\pm.007$ &
.336/.320/.336 & $.331\pm.009$ \\
\bottomrule
\end{tabular}
\caption{Final $k=32$ DAS results at layer 37 and the final pre-generation
position, across three independently initialized seeds. Standard deviations
are sample standard deviations across seeds.}
\label{tab:ministral-das}
\end{table}

The learned spaces are causally meaningful but not complete bottlenecks.
Mean IIA$_{\mathrm{AR}}$ ranges from $0.299$ to $0.401$ across the four
conditions: $0.299$ for $T_+$, $0.401$ for $T_-$, $0.305$ for $I_+$ and
$0.331$ for $I_-$. The effect is stable across independent DAS seeds, and
teacher-forced and autoregressive scores remain of the same order. We
therefore interpret the 32-dimensional space as a compact late numerical
subspace with moderate causal control, rather than as the unique or fully
sufficient causal state for the answer.

\subsection{Exact separation of the current digit readout}
\label{app:ministral-readout}

Let $U_D$ be an orthonormal basis for the centered unembedding directions of
the ten digit tokens. As in the primary model, centering removes the common
mean and leaves a rank-9 digit-discriminative span. For each learned
$R\in\mathbb{R}^{5120\times 32}$, we decompose
\[
    U_D^\top R = A\Sigma V^\top,
    \qquad
    \widetilde R = RV.
\]
This rotates coordinates only inside the learned DAS space. The overlap has
rank 9 for every condition and DAS seed, so we write
\[
    \widetilde R=[D\;L],
    \qquad
    \dim(D)=9,\quad \dim(L)=23,
\]
where $D$ carries all linear coupling of the DAS space to the centered digit
readout and $L$ lies in its null space.

The decomposition is numerically exact at the precision of the computation.
Across the twelve condition--seed spaces,
$\|U_D^\top L\|_F$ lies between approximately
$4.7\times10^{-9}$ and $5.4\times10^{-9}$,
$\|D^\top L\|_F$ between $2.1\times10^{-7}$ and
$3.6\times10^{-7}$, and
$\|DD^\top+LL^\top-RR^\top\|_F$ is approximately
$2.0\times10^{-7}$. Thus no part of the learned causal subspace is discarded.

The original DAS spaces are strongly enriched for current digit-readout
directions. With
\[
    E_D=\frac{\|U_D^\top R\|_F^2}{k},
\]
the mean overlap is $0.0830$, $0.0977$, $0.0935$ and $0.1082$ for
$T_+$, $T_-$, $I_+$ and $I_-$, respectively. A random 32-dimensional
subspace in a 5120-dimensional residual stream has expected overlap
$9/5120=0.00176$, making the observed values approximately $47$--$62$ times
larger than this expectation. This establishes strong coupling to all nine
independent digit contrasts, but does not imply that the complete digit span
is literally contained unchanged inside $R$.

\subsection{Shared numerical geometry in the readout-free component}
\label{app:ministral-geometry}

We next ask whether the 23-dimensional $L$ component remains numerically
organized once every direction linearly coupled to the current digit output
has been removed. All values $0,\ldots,99$ are represented in every
condition--seed space. For each of three deterministic value splits, 80 values
are used to fit the cross-condition relation and 20 values are held out
entirely. Scaled orthogonal Procrustes maps, together with the coordinate
offset needed to place value centroids, are fitted using only transitions whose
endpoints lie in the 80 training values. Evaluation uses only transitions
between the 20 unseen values. Across twelve directed condition relations, all
$3\times 3$ source/destination DAS-seed combinations and three value splits,
this gives $n=324$ pairwise evaluations.

\begin{table}[t]
\centering
\small
\begin{tabular}{lccc}
\toprule
Representation/control & Transition cosine & Top-1 & Top-5 \\
\midrule
Exact $L$ (23d) & $.8963\pm.0463$ & $93.35\%$ & $99.85\%$ \\
Random readout-null (23d), mean & $.6950$ & $72.30\%$ & $94.17\%$ \\
Label permutation, mean & $\approx 0.0000$ & $5.00\%$ & $25.01\%$ \\
\bottomrule
\end{tabular}
\caption{Held-out numerical geometry in the Ministral readout-free space.
Top-1 and top-5 retrieval are among 20 unseen candidate values, giving chance
levels of $5\%$ and $25\%$. The random row is the mean over 200 independent
dimension-matched readout-null projections; the permutation row is the mean
over 500 numerical-label permutations of the real $L$ spaces.}
\label{tab:ministral-geometry}
\end{table}

The held-out transition cosine is $0.8963\pm0.0463$, and the transported state
retrieves the correct unseen numerical value at $93.35\%$ top-1 and $99.85\%$
top-5. A fitting-free RSA control gives mean cross-condition Spearman
$\rho=0.8541\pm0.0898$. The corresponding pairwise heatmaps show no weak
relation: the high global score is not driven only by one easy pair, but is
present across modality changes, operation changes, and relations that change
both.

A single random readout-free projection also preserves substantial numerical
structure, so the geometry is not confined uniquely to the DAS-derived
remainder. We therefore evaluate a distribution of 200 independent random
23-dimensional readout-null subspaces using the same held-out protocol. Their
mean transition cosine is $0.69497\pm0.00759$ with 95\% interval
$[0.67998,0.70875]$; mean top-1 is $0.72304\pm0.01638$ with interval
$[0.69010,0.75141]$; mean top-5 is $0.94167\pm0.00669$ with interval
$[0.92807,0.95387]$; and mean RSA is $0.60480\pm0.02123$ with interval
$[0.56828,0.64370]$. None of the 200 random projections reaches the observed
$L$ score for any metric, giving the finite-sample resolution floor
$p_{\mathrm{rand}}=1/201$.

The numerical-label null is stronger still. Across 500 permutations, mean
transition cosine is approximately $-6\times10^{-6}$, top-1 is $0.04997$,
top-5 is $0.25010$, and RSA is $-0.00143$. None reaches the observed score,
so $p_{\mathrm{perm}}=1/501$ for all four metrics. The two controls together
support a narrower conclusion than uniqueness: numerical geometry is broadly
distributed in the readout-orthogonal residual representation, but the
readout-free component selected inside the causal DAS space is strongly
enriched for the shared numerical structure.

\subsection{The pairwise relations synchronize into one latent frame}
\label{app:ministral-sync}

Pairwise Procrustes fits could succeed while remaining mutually inconsistent.
We therefore repeat the synchronization analysis in the $L$ spaces. Matched
DAS seeds are used so that each graph contains one coherent set of four
condition-specific spaces. Across three seeds and three value splits, the
analysis gives $n=108$ directed relation evaluations.

\begin{table}[t]
\centering
\small
\begin{tabular}{lccc}
\toprule
Map & Transition cosine & Top-1 & Top-5 \\
\midrule
Direct pairwise & $.89642\pm.04611$ & $93.33\%$ & $99.82\%$ \\
Full synchronization & $.89482\pm.04515$ & $93.84\%$ & $99.86\%$ \\
LORO synchronization & $.89020\pm.04357$ & $93.89\%$ & $99.91\%$ \\
\bottomrule
\end{tabular}
\caption{Global consistency of the Ministral $L$ spaces. LORO removes both
directions of one complete condition relation before synchronization and
reconstructs the missing relation only through the remaining task graph.}
\label{tab:ministral-sync}
\end{table}

Replacing independently fitted pairwise maps by one synchronized frame changes
mean transition cosine by only $0.00160$. More importantly, when one complete
relation is withheld in both directions, LORO still reaches transition cosine
$0.89020$, an absolute drop of only $0.00623$ from direct fitting, or roughly
$0.7\%$ relative. Retrieval does not degrade. The four task-specific
readout-free spaces therefore behave as different coordinate systems of one
globally consistent numerical geometry: a relation whose rotation and scale are never fitted directly can be reconstructed from the remaining graph. This establishes
global consistency up to the fitted task-specific rotations, scales and
offsets; it does not imply a literally identical residual-space subspace.

\subsection{The readout-free identity causally redirects the continuation}
\label{app:ministral-continuation}

Geometry alone does not establish functional use. We therefore test whether
changing the numerical identity in $L$ affects future generation while holding
the currently required digit fixed. Let $y$ and $\widetilde y$ be two
two-digit donor values with the same first digit and different second digits,
for example $47$ and $43$. For a base activation $h_b$, decompose the DAS edit
into
\[
    \Delta h_D^{b\rightarrow y}=\Pi_D(h_y-h_b),
    \qquad
    \Delta h_L^{b\rightarrow y}=\Pi_L(h_y-h_b).
\]
The matched state is
\[
    h_{\mathrm{match}}
    = h_b+\Delta h_D^{b\rightarrow y}+\Delta h_L^{b\rightarrow y},
\]
whereas the latent mismatch keeps the readout-coupled component fixed and
changes only the numerical identity supplied through $L$,
\[
    h_{\mathrm{mis}}
    = h_b+\Delta h_D^{b\rightarrow y}
      +\Delta h_L^{b\rightarrow\widetilde y}.
\]
The experiment uses up to 128 deterministic triplets per task, all four tasks,
and all three DAS seeds.

At the current output step, the shared first-digit probability is $0.97876$ in
the matched state and $0.97130$ under the latent mismatch, a change of only
$-0.00746$. The intervention therefore does not simply replace the whole
answer at the current readout. At the following digit, however, the preference
reverses. Under the matched intervention,
$P(y_2)=0.33249$ and $P(\widetilde y_2)=0.05826$; after replacing only $L$,
these become $0.05945$ and $0.32315$. Greedy second-token choice shows the
same exchange: the selected $y_2$ rate falls from $0.38867$ to $0.05273$,
while the selected $\widetilde y_2$ rate rises from $0.05339$ to $0.38932$.
The effect propagates to full autoregressive generation: the matched state
generates toward $y$ at rate $0.28581$ and toward $\widetilde y$ at $0.05273$;
the mismatch instead generates toward $y$ at $0.04362$ and toward
$\widetilde y$ at $0.31315$.

\begin{table}[t]
\centering
\small
\begin{tabular}{lccccc}
\toprule
Task & Matched $P(y_2)$ & Mismatch $P(\widetilde y_2)$ &
Matched $\rightarrow y$ & Mismatch $\rightarrow\widetilde y$ &
$\Delta P(y_1)$ \\
\midrule
$I_+$ & .257 & .263 & .174 & .206 & $-.010$ \\
$I_-$ & .371 & .375 & .286 & .339 & $-.017$ \\
$T_+$ & .262 & .249 & .268 & .279 & $-.006$ \\
$T_-$ & .440 & .405 & .414 & .430 & $+.003$ \\
\bottomrule
\end{tabular}
\caption{Per-condition consistency of the same-first-digit latent mismatch,
averaged over the three DAS seeds. $\Delta P(y_1)$ is the mismatch-minus-matched
change in probability of the shared first digit. The qualitative continuation
effect appears in every modality and operation.}
\label{tab:ministral-continuation}
\end{table}

Controls separate value-specific continuation from generic perturbation.
A $D$-only intervention gives
$P(y_1)=0.89210$, $P(y_2)=0.07864$ and
$P(\widetilde y_2)=0.08089$, so $D$ alone does not strongly specify the
continuation. A random latent donor gives $0.93427$, $0.08079$ and $0.08408$,
respectively. Finally, a value-specificity control using an unrelated latent
donor compatible with the same first digit preserves
$P(y_1)=0.97482$ but gives only $P(y_2)=0.05391$ and
$P(\widetilde y_2)=0.05941$. First-digit compatibility is therefore
insufficient: the redirected continuation depends on the numerical identity
inserted through $L$.

The causal conclusion is temporal rather than digit-wise. The data do not show
that $D$ ``stores the first digit'' or that $L$ is a dedicated second-digit
subspace. Instead, holding the readout-coupled component fixed preserves the
immediately required digit, while changing numerical identity in the
readout-orthogonal component selectively redirects the subsequent digit and
the final autoregressive answer. The numerical identity in $L$ is therefore
not only geometrically organized; it is functionally involved in future
generation.

\subsection{Summary}
\label{app:ministral-summary}

Ministral reproduces the central mechanism despite a lower absolute
full-answer DAS intervention score. A compact late causal numerical space can
be found and separated exactly into a rank-9 component coupled to the current
digit readout and a 23-dimensional readout-null component. The latter retains
strong held-out numerical geometry across modality and operation
(transition cosine $0.896$, top-1 $93.35\%$, RSA $0.854$), is substantially
more structured than a distribution of random readout-null projections,
synchronizes into one common latent frame with negligible loss, and
reconstructs a relation never fitted directly. Most importantly, replacing
only its numerical identity leaves the immediate shared digit almost
unchanged while reversing the next-digit preference and redirecting the
freely generated answer. The replication therefore separates two questions
that need not coincide: how completely a small subspace controls the full
autoregressive output, and how strongly the underlying numerical state is
organized and functionally used.

\end{document}